\documentclass{article}

     \PassOptionsToPackage{numbers, compress}{natbib}

\usepackage[preprint]{neurips_2025}

\usepackage[utf8]{inputenc} 
\usepackage[T1]{fontenc}    
\usepackage{hyperref}       
\usepackage{url}            
\usepackage{booktabs}       
\usepackage{amsfonts}       
\usepackage{nicefrac}       
\usepackage{microtype}      
\usepackage{xcolor}         
\usepackage{graphicx}
\usepackage{amsmath}
\usepackage{wrapfig}
\usepackage{subcaption}
\usepackage{todonotes}
\usepackage{float}
\usepackage{multirow} 
\usepackage{ragged2e}
\title{Distinct Computations Emerge From Compositional Curricula in In-Context Learning}

\author{Jin Hwa Lee\textsuperscript{1},~Andrew K. Lampinen\textsuperscript{2},~Aaditya K. Singh\textsuperscript{1*},~Andrew M. Saxe\textsuperscript{1,3*} \\
\textsuperscript{1}University College London,
London, UK\\
\textsuperscript{2}Google Deepmind, Mountain View, CA, USA\\
\textsuperscript{3}CIFAR Azrieli Global Scholar, CIFAR\\
\textsuperscript{*}Co-senior authors\\
Correspondence: \texttt{jin.lee.22@ucl.ac.uk} \\
}

\begin{document}

\maketitle

\begin{abstract}
In-context learning (ICL) research often considers learning a function in-context through a uniform sample of input-output pairs. Here, we investigate how presenting a compositional subtask curriculum in context may alter the computations a transformer learns.
We design a compositional algorithmic task based on the modular exponential---a double exponential task composed of two single exponential subtasks---and train transformer models to learn the task in-context. We compare (a) models trained using an in-context curriculum consisting of single exponential subtasks and, (b) models trained directly on the double exponential task without such a curriculum. We show that models trained with a subtask curriculum can perform zero-shot inference on unseen compositional tasks and are more robust given the same context length. We study how the task and subtasks are represented across the two training regimes. 
We find that the models employ diverse strategies modulated by the specific curriculum design.
\end{abstract}

\section{Introduction}

Many complex real-world tasks require composing multiple constituent functions or subtasks. This notion of systematic compositionality has been extensively studied~\citep{Chomsky1999DerivationBP, Frege1948-FREUSU-3, Szabó_2024} and is argued to be a key feature of flexible intelligence, enabling ``infinite use from finite means.'' However, it has been a central controversy whether neural networks can exhibit human-like compositionality \cite{fodor1988connectionism,smolensky1988proper,lake2018generalization,lake2023human}. The recent success of Large Language Models (LLMs) has only brought this controversy to a head, given their often inscrutable nature yet remarkable generalization capabilities~\cite{brown2020language, wei2022emergent}.

A body of literature has studied the capabilities and limitations of compositional generalization in LLMs, including multi-hop reasoning~\citep{press2022measuring, yang2024largelanguagemodelslatently}, chain-of-thought reasoning~\citep{wei2022chain, 52454}, and the scratch-pad~\citep{nye2021show}. For example, in a multi-hop reasoning query such as ``the mother of the singer of ‘Superstition’'', LLMs seem to latently encode intermediate information ``Stevie Wonder is the singer of `Superstition''' and use it to produce the final answer~\citep{yang2024largelanguagemodelslatently}. Furthermore, LLMs can improve their performance on complex hierarchical tasks by decomposing the tasks into component subtasks~\citep{zelikman2023parsel,zhou2023least}. On the other hand, such compositional generalization in LLMs is not always guaranteed. For tasks defined by composition of several functions or subtasks, models often find a surrogate strategy rather than learning the underlying true compositional structure~\cite{dziri2024faith,kobayashi2024can}. 

To this end, it is important to understand the circumstances under which compositional generalization abilities may emerge, especially when the subtask or component knowledge is available. We approach this question from a data distributional perspective. While previous work showed that language-like data distributional properties such as skewedness and burstiness trigger in-context learning ability \cite{chan2022data}, another distinct property of natural language corpus is that it encompasses a variety of compositional structures. For example, many essays and informative writings contain multiple paragraphs, each of which introduces supporting arguments, and at the end follows a conclusion paragraph that composes all of the previous points. Texts of instructions or math problem solutions similarly contain compositional curriculum-like structure: first presents component knowledge, followed by the composition of these components. Inspired by this, we demonstrate that an in-context curriculum data structure ---in-context examples of components and their composition--- can give rise to robust compositional generalization ability. 

\begin{figure}
\begin{center}
\vspace{-0.25cm}
\includegraphics[width=\textwidth]{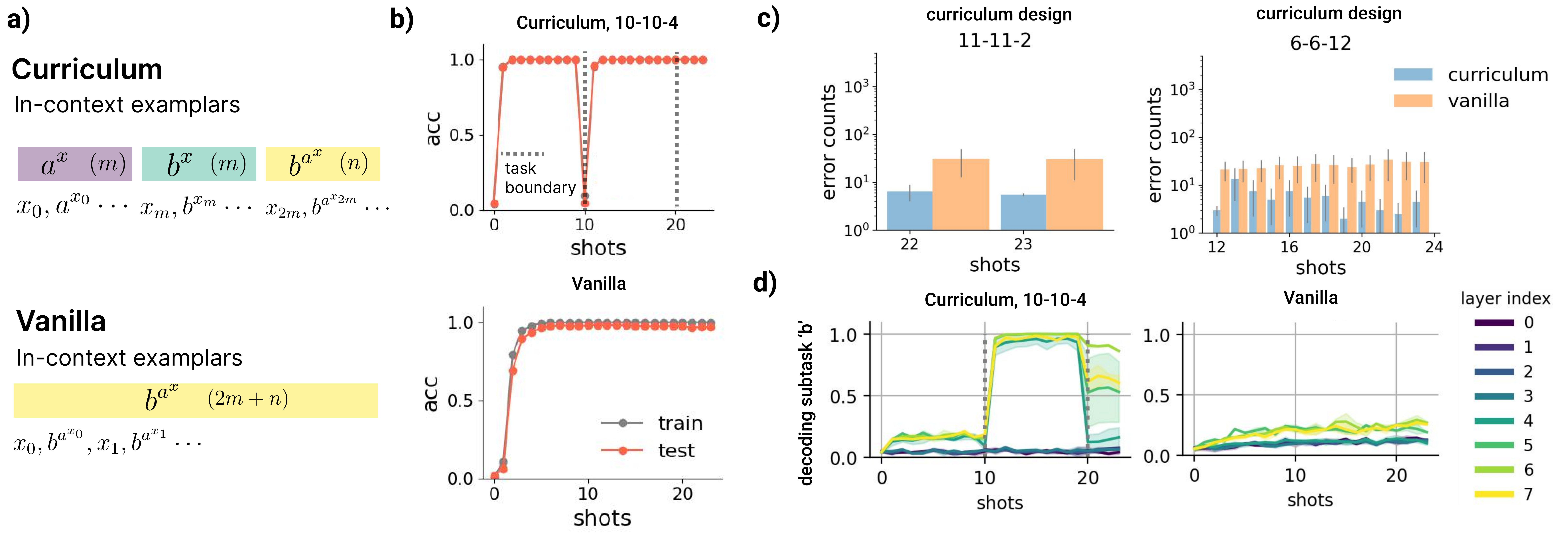}
\end{center}
\caption{Overview of the setup and key results. \textbf{a)} Task schema. In curriculum training, each training sequence is composed of $m$ exemplars for two single-exponential tasks defined by $a$ and $b$, respectively, followed by $n$ composite double-exponential task exemplars. In vanilla training, the model is trained with a sequence of $2m+n$ in-context exemplars for the double-exponential task defined by task parameters $(a,b)$. \textbf{b)} Example asymptotic accuracy at the end of training on training and evaluation tasks for curriculum training ($m=10$, $n=4$) (top) and vanilla training (bottom). \textbf{c)} Comparison of the error counts for the last compositional block for curriculum training and vanilla training (left: $m=11, n=2$, right: $m=6, n=12$). The curriculum condition (blue) enhances robustness on unseen compositional task block compare to vanilla training (orange).  \textbf{d)} In-context curricula (as shown in (a) top) promote representation of intermediate task parameters. We show linear probe decoding accuracy of task parameter $b$ from unseen evaluation sequences. The curriculum-trained model represents the task parameter in compositional task block (last 4 shots) while the vanilla-trained model does not.}
\label{figure1}
\vspace{-0.5cm}
\end{figure}

Concretely, we design an algorithmic task utilizing the composition of two modular exponential tasks defined by two exponent bases, $(a, b)$. To emulate in-context compositional curriculum structure, we design subtask curriculum with examples of single exponentials from each base followed by compositional double exponentials (see Figure~\ref{figure1} a). We train the model with example sequences sampled from sets of $(a,b)$ and evaluate its generalization ability on unseen combinations of $(a, b)$.
We first demonstrate that a model trained with subtask curriculum is capable of zero-shot inference on unseen compositional task queries and shows higher robustness compared to a model trained without a curriculum (Section~\ref{results_robustness}). We show evidence that the curriculum enables the model to represent and combine the task parameters to be composed (Section~\ref{results_representation}). Finally, we study how the length of the curriculum affects the model's learning and potentially which strategy, or even mixture of strategies, is used to solve the compositional task (Section~\ref{results_which}-\ref{results_when}).

Our results highlight that ``in-context curricula'' can provide higher-level correlational structure between subtasks and compositional task in-context that lead to models employing a more compositional form of in-context learning at test time. Furthermore, the degree to which these correlations are present in an input context modulates the trade-off between compositional and standard few-shot in-context learning strategies.

\section{Experimental Setup}
\label{experimental_setup}

\subsection{Task}

We use a modular arithmetic task of composition of an exponential function, namely $b^{a^x}\mod P$, referred to as the modular double exponential task. Inspired by well-studied linear modular arithmetic tasks such as summation, multiplication or both \cite{he2024learning, nanda2023progressmeasuresgrokkingmechanistic}, we chose the modular double exponential function for its greater complexity while still offering a deterministic functional mapping and effectively constraining the vocabulary size. As shown in Figure~\ref{figure1} a, we design a \textit{curriculum} of in-context exemplars, which provides blocked examples for $y=a^x \mod P $ and  $y=b^x \mod P$, followed by $y=b^{a^x} \mod P$. In contrast, \textit{vanilla} in-context exemplars consist only of $y=b^{a^x} \mod P$. We train transformer architecture on these sequences using a next token prediction task for every $(x,y)$ pair in the sequence, rather than only on the final query. 

To make a fair comparison, we provide single exponential task exemplars in \textit{vanilla} training as well, but the main difference is that they never appear in the same sequence as exemplars for the compositional task. Curriculum- and vanilla-training thus see the same set of task inputs and outputs at an exemplar level, with the key difference being in-context correlations at the \textit{sequence level}: curriculum-training see both subtask and compositional task in-context, while in vanilla training each sequence is just from one task.

In every example sequence, we randomly sample task parameters $(a,b)$, and the model needs to learn to adapt its answer in-context according to $(a,b)$. The task combinations $(a,b)$ seen during the training include all possible individual $a$, and $b$, but not all pairs. We evaluate the trained model on unseen combinations of $(a,b)$. We used $80/20$ split of all possible pairs for train and test. We permit integers $x \in [0,P)$, and $a$ and $b$ are sampled from the primitive roots of $P$. Throughout the main experiments, we focus on $P= 59$, and we extend our findings to other values of $P$ in Appendix~\ref{app:other_modulo_robustness}.

In the \textit{curriculum} setting, we use an equal curriculum length $m$ for each single exponential task and $n$ for the compositional task, leading to total length $2m + n$. We use the same $2m + n$ exemplars of compositional task for the \textit{vanilla} setting. While varying the compositional task length in the curriculum, we maintain the importance of the compositional task equal to each single exponential task in the curriculum by controlling the weighting factor for the loss contribution from the compositional task (namely, making it such that the loss from the compositional task is 1/3 of the total loss). Similarly, for a fair comparison, in the vanilla setting the network sees sequences for a single exponential task as well, with a ratio of 2 to 1 (to match the overall weight of the compositional task to the curriculum setting). By doing this, we effectively make it so that the same (exemplar, label) pairs are seen in both curriculum and vanilla settings, with the same loss weight for single vs. double exponential tasks. The key difference between the two settings is the \textit{in-context correlations}: in the curriculum setting, these correlations are more complex/hierarchical (possibly leading to in-context compositional learning), while in the vanilla setting, they focus on single function learning (the standard few-shot ICL setting).

We fix the total context length to $48$, corresponding to $24$ pairs of $(x,y)$, and ensure that all $x$ for each task sequence in context are unique.

\subsection{Model and Training}
\begin{wrapfigure}{r}{0.4\linewidth}  
    \vspace{-1.0cm}  
    \begin{center}
        \includegraphics[width=\linewidth]{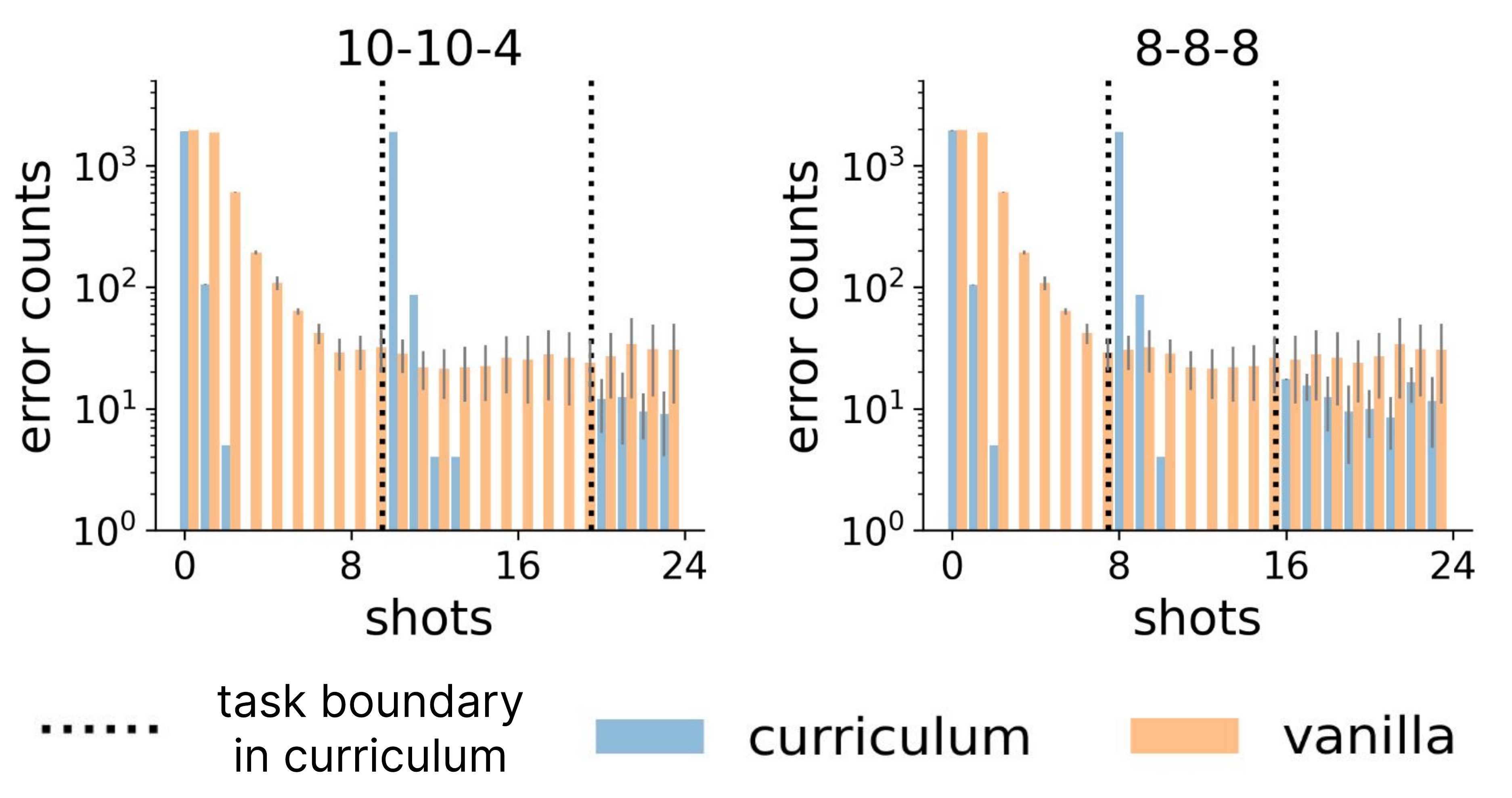}
    \end{center}
    \vspace{-5pt}  
    \caption{Errors across entire context in vanilla vs. curriculum model. Gray dotted lines indicate the task boundaries in the curriculum. The curriculum model generalizes to each subtask after few-shots (first two single exponential tasks) and once the compositional task is present, fewer errors occur compare to vanilla setting from the zero-shot on (after the second gray dotted line). }
    \vspace{-1.5cm}
    \label{figure_errors_all}
\end{wrapfigure}
We train 8-layer transformers with sinusoidal positional embeddings with time constant of 120, a hidden dimension size of 128, and 8 heads, using the Adam optimizer, a learning rate of $7.5 \times 10^{-4}$, and a batch size of 512. All results we report are based on 2 data seeds and trained with $2 \times 10^8$ sequences unless otherwise mentioned. Specifically, for the vanilla model, we further trained it up to $3 \times 10^8$ sequences to ensure that the model's performance is saturated.
See Appendix~\ref{app:example_loss} for example loss curve and performance evolution.

\section{Results}
\label{results}

\subsection{In-context curricula increase the robustness of in-context learning of a complex compositional task}
\label{results_robustness}

First, we demonstrate that the subtask curricula can make in-context learning of unseen compositional task more robust. As we see in Figure ~\ref{figure1} b, both the vanilla and curricula settings tend to achieve high accuracy. Here, we closely examine the performance difference through analyzing the error counts after each number of in-context exemplars for 2K evaluation sequences sampled with unseen task parameter combinations $(a,b)$ averaged over 2 seeds. In Figure~\ref{figure_errors_all}, we show the trend of error counts across the entire context comparing the vanilla (orange) and curriculum (blue) settings. First, we see that the curriculum model first generalizes to the two subtasks (first two blocks) after few-shots. We notice that the curriculum enables zero-shot inference on the compositional task (after the second dotted task boundary). Compared to the non-informative zero-shot inference of the vanilla model, the zero-shot performance on the compositional task exhibits significantly fewer errors in the curriculum model. In contrast, the vanilla model results in a gradual decrease of errors as more examplars are shown, but the errors eventually saturate. 

\begin{figure}

    \begin{center}
    \includegraphics[width=\linewidth]{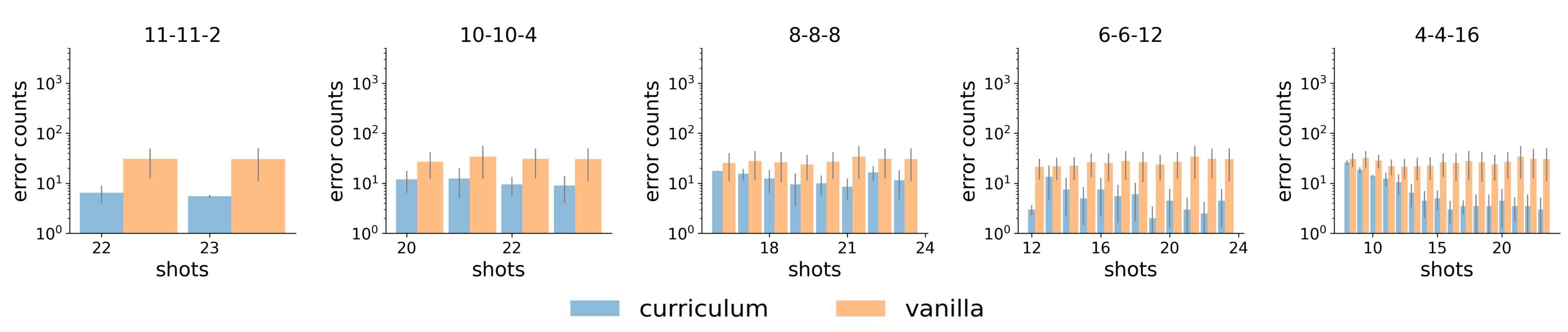}
    \end{center}
    \vspace{-0.2cm}
    \caption{In-context error counts of \textit{compositional task} in vanilla model vs. curriculum models. We zoom into the compositional task block of Figure ~\ref{figure_errors_all}. The leftmost point is zero-shot of compositional task for  curriculum model. We compare the error counts on compositional task of curriculum (m-m-n) models to that of corresponding $n$ last exemplars from the vanilla model. The curriculum model shows higher robust (fewer error) than the vanilla model. See Appendix~\ref{app:error_viz} for more visualizations.}
    \vspace{-0.5cm}
    \label{figure_robustness}
\end{figure}

In Figure~\ref{figure_robustness}, we focus on the generalization ability on the compositional task by comparing the errors on $n$ compositional task exemplars in each curriculum setting and the corresponding $n$ last exemplars in the vanilla setting. We observe that the curriculum models result in higher robustness (fewer errors) compared to the vanilla model. Note that most of the curriculum settings exhibit rather consistent errors throughout the compositional task, but the rightmost curriculum (4-4-16) shows further decreasing of errors with more compositional exemplars without getting more subtask information. This suggests the model utilizes in-context compositional examples rather than solely relying on subtask curriculum.  We dive deeper into this aspect in Section~\ref{results_which}-~\ref{results_when}.

Based on the above behavioral evidence that in-context curriculum can make the compositional generalization more robust, we ask what is the mechanism behind this. 

\subsection{In-context curricula promote representation of compositional subtasks}
\label{results_representation}
\begin{figure}
\begin{center}

\includegraphics[width=\linewidth]{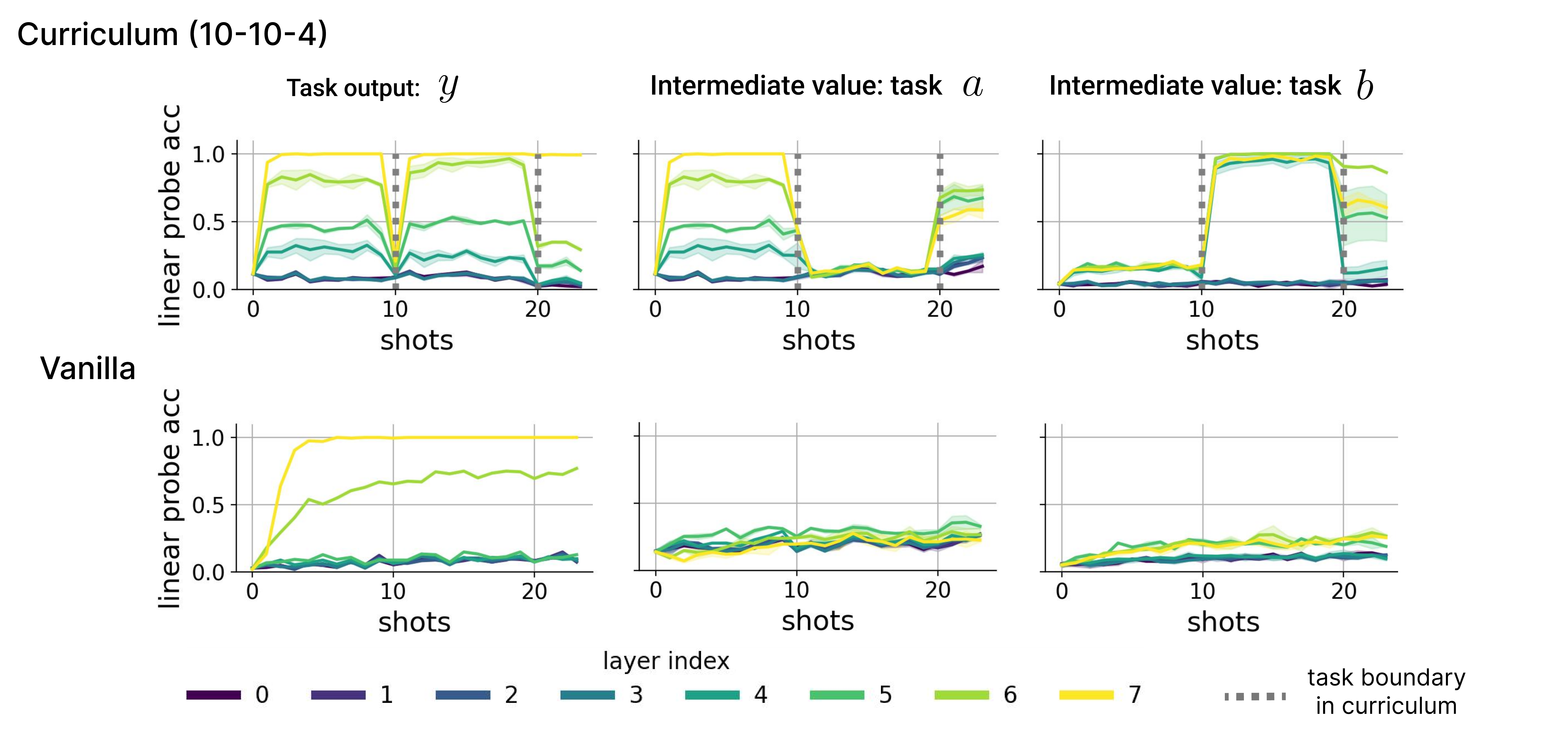}
\end{center}
\vspace{-0.2cm}
\caption{Linear probe decoding accuracy of target $y$ for corresponding tasks and intermediate values from the subtasks needed for compositional task in the vanilla (top) and curriculum (10-10-4) setting (bottom). Both settings show high decoding of $y$ for corresponding task blocks. In the curriculum setting, the intermediate values from the subtasks $(a,b)$ required for compositional computation show high decodability in corresponding single task block and importantly, in compositional task block (shot 20-23), but not in the vanilla model. Noticeably, in the curriculum model, the highest decodability of intermediate values in compositional task block come from not the final layer but earlier layers (light green) indicating layer-wise processing of compositional computation. } 
\vspace{-0.5cm}
\label{figure_probe}
\end{figure}

In the previous section, we observed that the models can benefit from having subtask curricula for solving compositional tasks in-context, particularly its zero-shot inference ability on the compositional task and higher robustness compared to the vanilla model. In this section, we ask \textit{how} in-context curricula may enable this compositional solution. We hypothesize that the in-context curriculum of the single exponential tasks provides unambiguous information about the subtasks that constitute the compositional task, thereby facilitating the model’s learning of a compositional solution to the double exponential task.

To test this hypothesis, we investigated how the model represents intermediate values from constituent subtasks required for compositional solution. We trained a linear classifier\footnote{While simple, linear probing is the typical first pass at studying internal representations \cite{gurnee2023finding, nanda2023emergent}.} with the hidden representation of each layer from evaluation sequences to decode the intermediate values required for compositional computation---$a^x \mod (P-1)$ and task parameter $b$\footnote{$b^{a^x} \mod P = b^{(a^x)\mod(P-1)}\mod P$.}--- in compositioanl task block. We additionally probe for corresponding task output $y$ at each position and the subtask information in the subtask blocks (task $a$ computation and task parameter $b$). We used $80/20$ split of the 1K unseen test sequences for the linear probe training and testing of the decoding accuracy \footnote{See Appendix ~\ref{app:linear_probe_baseline} for control experiment with shuffled labels to confirm the baseline probe performance.}. See Appendix~\ref{app:linear_probe_diff_curr_len} for more results in other curriculum designs and experimental details. 

In Figure~\ref{figure_probe}, we first observe that the decoding of $y$ values of corresponding tasks becomes near perfect at the final layer, as expected from the overall high accuracy in both the vanilla and curriculum models. However, we find a noticeable difference in decoding of intermediate computation values. 

With in-context curricula, each subtask's information is well decoded in the corresponding subtask block. Most of all, the intermediate computation values involving the subtasks $(a,b)$ are highly decodable in compositional task block (shot 20-23), suggesting that the subtask representation inferred from the curriculum is utilized in the compositional task. Especially, the high decoding accuracy at the zero-shot of compositional task (shot 20) indicates that the model readily utilizes the intermediate computation values inferred from the subtask curriculum to solve compositional task. In contrast, the vanilla-trained model shows lower decoding accuracy of intermediate values required for compositional computation, indicating that training with in-context curricula is what incentivizes the more compositional representations (at least, in terms of linear decodability). 
\begin{wrapfigure}{r}{0.5\linewidth}  
    \begin{center}
        \includegraphics[width=\linewidth]{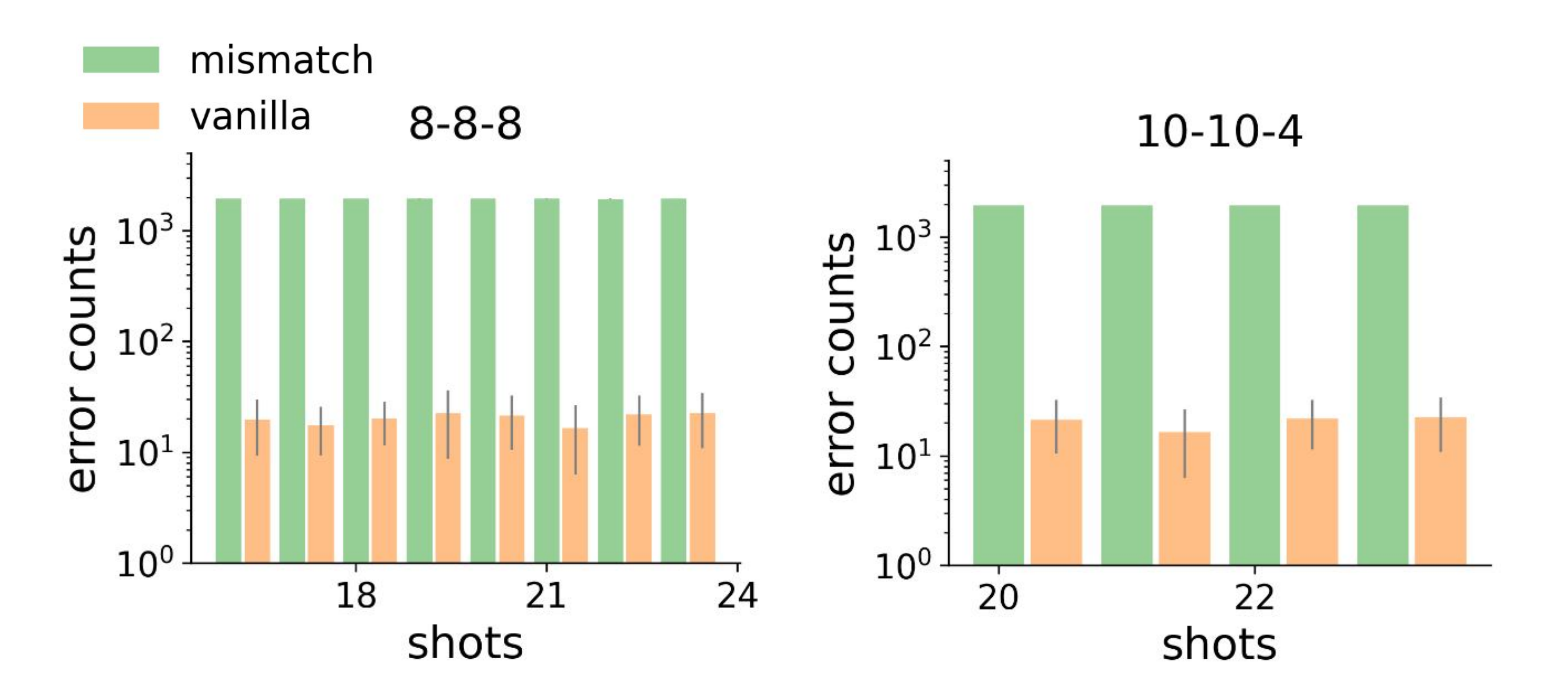}
    \end{center}
    \vspace{-5pt}  
    \caption{The curriculum model performance on mismatch sequences:single exponential tasks of parameter $(a,b)$ followed by double exponential task with mismatching paramters $(a',b')$. The models trained with curriculum (8-8-8) and (10-10-4) fail on the mismatching compositional task, showing no improvement with few-shot compositional examples. It indicates that the model relies on composition of subtask information given in the curriculum to solve the compositional task and thus cannot solve the problem without correct subtasks information.}
    \vspace{-0.7cm}
    \label{figure_mismatch1}
\end{wrapfigure}

Next, we take a closer look at the layer-wise decoding accuracy. In the curriculum model, we observe that the highest decoding of the intermediate computation values in the compositional task are not achieved in the last layer but in the earlier layers (layer 5-6). This suggests the layer-wise processing of subtask information in compositional block. That is, the representation from the subtask curriculum blocks in the context is transferred and processed in the earlier layers and go through further computation to combine those to perform compositional task. Additionally, we visualize the attention pattern of the heads in the earlier layers where we can find heads that attend to the earlier curriculum block from the compositional task block in Appendix \ref{app:attention}. Collectively, these results indicate that the curriculum-trained model encodes and utilizes the intermediate values required for the compositional task inferred from the curriculum.

To provide further evidence for our hypothesis, we test the curriculum model with mismatch sequences -- single exponential tasks with task parameter $(a,b)$ followed by double exponential task with $(a', b')$, where the compositional task block uses mismatching task parameters than the subtasks shown in context. If the model is using compositional strategy, we expect the model to fail on this task, as the subtask representations are not informative. In Figure \ref{figure_mismatch1}, we show that this prediction is consistent in the curriculum model trained with short compositional task block (10-10-4, 8-8-8, etc, see Appendix ~\ref{app:mismatch} for more results).

In brief, we show evidence that the model trained with in-context curriculum encodes the subtask information in its internal representation and is capable of using them for the compositional task using linear probes and controlled experiment. We show the decoding of intermediate values, which suggests step-by-step compositional computation in the curriculum model. In contrast, the vanilla model does not necessarily represent such intermediate computation from the subtasks. The mismatch experiment further confirms that having incorrect subtask information causes the curriculum-trained model to fail on the compositional task.

\subsection{In-context curriculum designs change the model's strategy on a compositional task} \label{results_which}

\begin{wrapfigure}{r}{0.3\linewidth}  
    \vspace{-1cm}  
    \begin{center}
        \includegraphics[width=\linewidth]{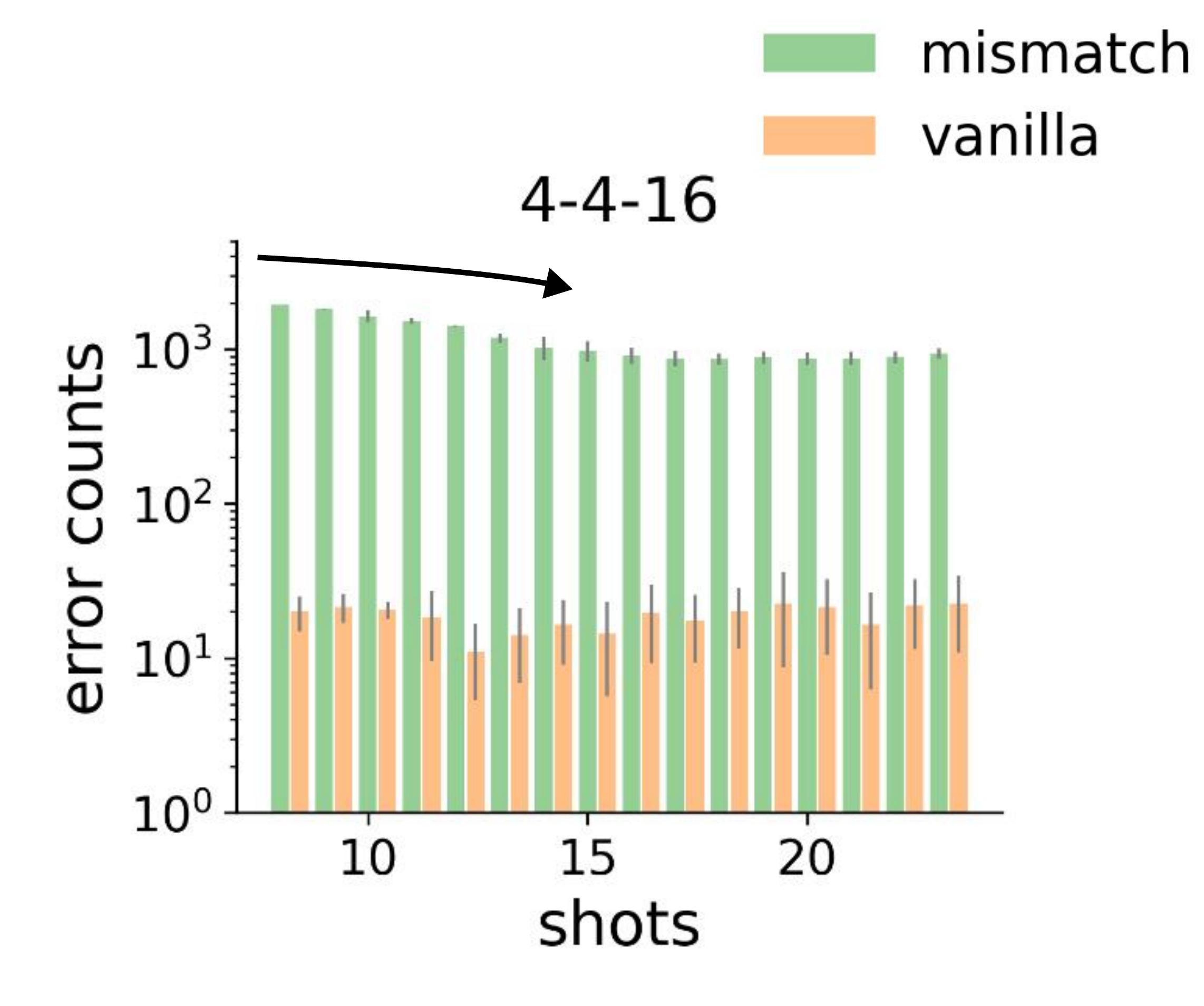}
    \end{center}
    \vspace{-15pt}  
    \caption{In mismatch experiment, the model trained with longer compositional context shows slight decreasing of error suggesting the model does not only relies on compositional strategy but vanilla few-shot learning strategy as well.}
    \label{figure_mismatch_short}
    \vspace{-0.5cm}
\end{wrapfigure}

We demonstrated that the in-context curriculum can enhance the model's robustness on compositional generalization possibly through promoting the encoding of the relevant subtasks representation. However, since the compositional task can be learned fairly well without curriculum but with standard few-shot in-context learning (as seen in the vanilla model), it is unclear why the model learns to use the subtask information inferred from the single exponential task examples. In other words, the model can learn the compositional task independently, even when the curriculum sequence is provided, without relying on the subtask information. Our earlier comment on the trend of decreasing error within the compositional task block in the curriculum (4-4-16) reflects this possibility (Figure~\ref{figure_robustness}).

In the mismatch experiment, we can find a hint of this behavior in the model trained with longer compositional task block. In Figure \ref{figure_mismatch_short}, we see slight performance improvement despite mismatch of task parameters when the compositional task length is longer (4-4-16). This is not only due to the more examples given since the (10-10-4) or (8-8-8) models completely fail given 4-8 examples (Figure~\ref{figure_mismatch1}), but the (4-4-16) model shows improvement already from one-shot example. This implies that the model can still make some inference without compositional strategy, suggesting the model employs possibly both strategies ---compositional and vanilla strategy---.

\begin{figure}
\begin{center}
\includegraphics[width=\linewidth]{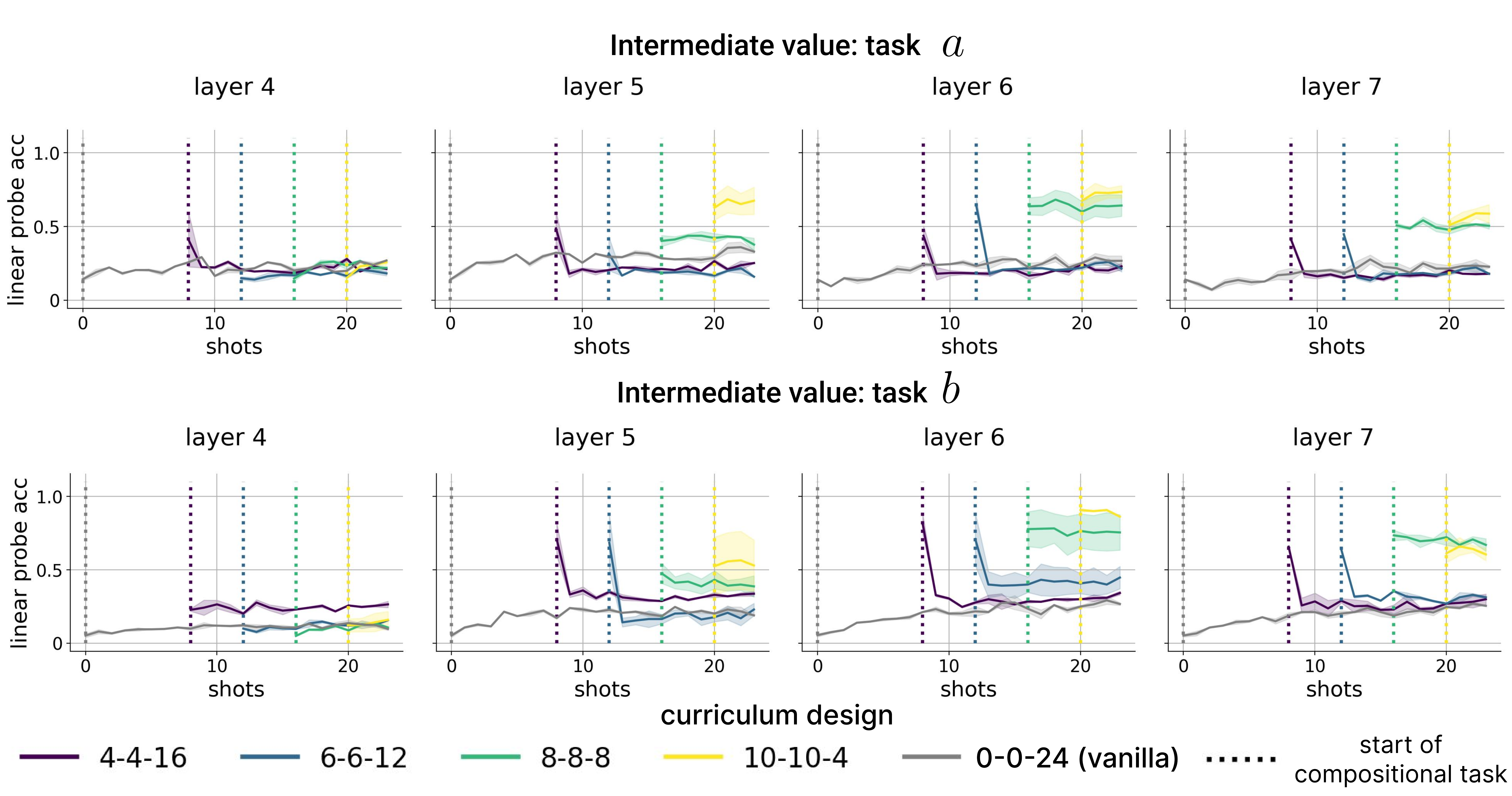}
\end{center}
\caption{Evidence of mixed strategy between vanilla and compositional solution modulated by curriculum design. We show decodability of intermediate computation involving each subtask $(a,b)$ in the compositional task block across different curriculum designs in layer 4-6. At the zero-shot of the compositional task (at the dotted lines), both intermediate values are highly decodable in longer compositional task curriculum design (6-6-12, 4-4-16), and it decreases afterwards, indicating existence of both vanilla and compositional strategy. For the shorter compositional task design (10-10-4, 8-8-8), we observe high decodability consistently in entire block but the level of decodability is relatively higher with shorter compositional task length (yellow vs. green). These results suggest complex interplay between the different curricula design and a strategy the model employs. }
\vspace{-0.5cm}
\label{figure_strategy}
\end{figure}

This sign of mixed strategies becomes more clear in the linear probes of subtasks intermediate computation. In Figure~\ref{figure_strategy}, we show linear probing of compositional task block from different curriculum designs. In the curriculum design with long compositional task block (6-6-12, 4-4-16), we observe the intermediate values show high decodability at the zero-shot (at the dotted lines) of the compositional task, but kept low value similar to the vanilla setting in the rest. This can be interpreted that the model employs a compositional strategy exploiting subtask information on the zero-shot of compositional task since the vanilla strategy cannot provide any meaningful inference on the zero-shot example, while the model still utilizes vanilla few-shot learning strategy as well. On the other hand, the curriculum setting with a shorter compositional task shows consistently high decodability of intermediate computations across the entire block, indicating that the model is predominanlty relies on a compositional strategy.

It is notable that the intermediate value decodability is not all-or-none for (8-8-8) or (10-10-4). Rather, it increases continuously as the compositional task length decreases (see green and yellow lines in Figure~\ref{figure_strategy}), reflecting that the compositional strategy is not binary but graded. That is, the possible choice of strategy that a model employs is not mere binary choice of either compositional or non-compositional, but rather lies on a continuous spectrum. 

\subsection{In-context curricula designs modulate order in which the tasks are learned}
\label{results_when}

\begin{figure}
\begin{center}
\includegraphics[width=0.9\linewidth]{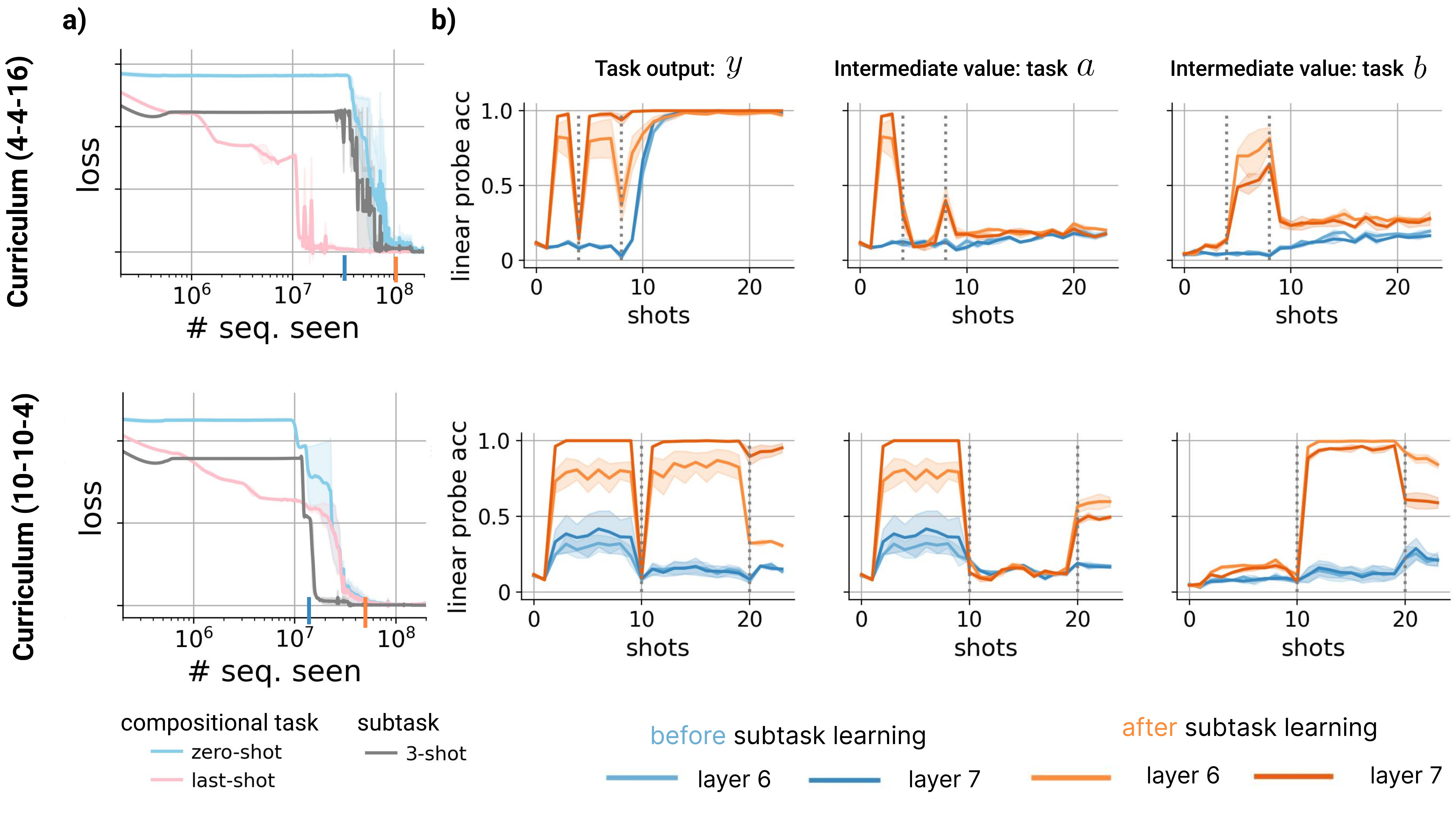}
\end{center}
\vspace{-0.05in}
\caption{\textbf{a)} Top: In the curriculum (4-4-16), the model can solve compositional task after many examples before subtask learning (pink before gray). The zero-shot loss decreases sharply only after the subtask lerning (blue after gray). Bottom: In the curriculm (10-10-4), the subtask learning is followed by the compositional task learning. The learning of both zero-shot and the last shot happen almost spontaneously (pink and blue after gray). \textbf{b)} Linear probe at before and after the subtask learning (blue and orange mark in panel a). Top: In the curriculum (4-4-16), we see that the model readily make good prediction on $y$ in compositional task block after few in-context examples without learning subtasks (blue, near-1.0 accuracy after 12 shots). After the subtasks are learned (orange), the model is capable of zero-shot inference where we can find representation of the subtasks (note the increased decodability of $y$ at shots 8-10, and corresponding peaks in decodability of intermediate values from $a$ and $b$). Bottom: In the curriculum (10-10-4), the model can generalize to the compositional task only after the subtasks are learned, as evidenced by low decodability of $y$ before subtask learning (blue) and the high decodability of $y$ and intermediate values after subtask learning (orange). We also see here that subtasks representations are more decodable in earlier layers. See Appendix~\ref{app:probe_before_after} for results on other layers.}
\vspace{-0.7cm}
\label{figure_before_after}
\end{figure}

Next, we look into the loss evolution and linear probe across the training phase to understand how do the different strategy choices develop. In Figure~\ref{figure_before_after} a, we observe a clear difference in the order in which each task is learned when the curriculum length is varied. With long compositional task sequence (4-4-16), the model is capable of vanilla few-shot learning on compositional task before the subtasks being learned (pink before gray). Nevertheless, the zero-shot loss of the compositional task decreases substantially shortly after the subtask learning (blue after gray). On the other hand, with shorter compositional task length (10-10-4), the rapid learning of compositional task happens only after the subtask learning (pink and blue after gray). Furthermore, both the zero-shot and the last-shot loss decrease almost simultaneously, suggesting the subtask information is utilized in the entire compositional task block. These observations show that different curricula can change the training dynamics, that is, in which the order of the tasks are learned.

We find that development of the subtask representation in different curriculum settings is also aligned with the above observation. In Figure~\ref{figure_before_after} b, we show linear probe of the model checkpoints before and after the subtask learning (marked with the colored bars in Figure~\ref{figure_before_after} a). In curriculum (4-4-16), the model can readily solve compositional task before learning the subtasks with sufficient examples and its linear probe shows low decodability of intermediate subtask computation (suggesting vanilla few-shot learning strategy). The model becomes capable of zero-shot inference shortly after the subtask learning, facilitated by the subtask representation (increased decoding accuracy of intermediate subtask values at the zero-shot). On the other hand, we see that in the curriculum (10-10-4) the compositional task performance rapidly increases only after the subtask learning and entire compositional task block encodes the intermediate values from the constituent subtasks. 

Collectively, these suggest varying the correlational structure between subtask and compositional task given in-context by controlling curriculum design influences which task to be learned first and thus influencing the strategy that the model employs, in this case, compositional computation. 

\section{Related Works}
\label{related_works}
\textbf{In-context learning} has brought significant interest recently, particularly due to the emergent capabilities of LLMs~\citep{wei2022emergent}.
The in-context few-shot learning ability of language models \citep{brown2020language} can be seen as an instance of meta-learned few-shot learning, where the model adapts and generalizes to unseen input examples without requiring gradient updates or explicit meta-training---in contrast to earlier works on meta-learned few-shot learning~\citep{santoro2016meta, vinyals2016matching, wang2016learning}. Many studies~\citep{chan2022data,xie2022explanation,raventos2024pretraining} have highlighted the importance of data properties for in-context learning. A few studies ~\citep{hendel2023context, todd2023function} have explored how different in-context tasks can be represented in LLMs in the form of task vectors. \citet{russin2024human} show that ICL can match human patterns of compositional and non-compositional behavior in multi-output categorization tasks. 

\textbf{Compositionality in neural networks} has been a central controversy. However, the recent success of LLMs, and their language comprehension and generation---which are thought to be a hallmark of compositional ability of humans---may have altered the question into how and when compositionality can emerge, rather than its existence.
Indeed, various works have studied how meta-learning can enable systematic compositional generalization in neural networks \citep{mccoy2020universal,lake2023human}.
Correspondingly, evidence of compositional representation and computation in language models has been studied extensively in scopes ranging from representational structure \citep{tenney2019you,soulos2020discovering} to geometric manifolds \cite{lee2024geometric} to circuit level mechanistic interpretability \cite{geva2023dissecting, yang2024largelanguagemodelslatently, merullo2023circuit,todd2023function}.

\textbf{Curriculum learning} is critical in learning of humans and animals, well-attested in a body of  literature~\citep{skinner2019behavior,elio1984effects,clerkin2017real,dekker2022curriculum}. While its potential importance has long been acknowledged in machine learning community~\citep{bengio2009curriculum, wang2021survey}, the benefit from curriculum has been shown marginal in standard supervised learning benchmarks~\citep{wu2021when}. However, the right curricula show greater significance in the context of reinforcement learning~\citep{karpathy2012curriculum, tessler2017deep,narvekar2020curriculum}. In particular, ~\cite{lee2024animals} shows theoretical evidence of importance of subtask curricula in learning compositional task. However, there has been little exploration of curricula within in-context learning.

\textbf{Modular arithmetic tasks} have been used in a rich body of literature to understand how sequence models, such as transformers, can implement the internal mechanisms required to solve these tasks. For example, \cite{grokking, zhong2023clockpizzastoriesmechanistic} used simple modular addition tasks to investigate the grokking phenomenon and demonstrate that transformers can implement multiple solutions. \citet{he2024learning} studied how transformers can learn skill composition in-context with out-of-distribution tasks. Our work builds on these findings by exploring how transformers can utilize a curriculum of subtasks provided in-context to achieve compositional generalization on more complex modular arithmetic tasks.

\section{Discussion}
\label{discussion}

We investigated how transformers can leverage inferred subtask information from an in-context curriculum to generalize to unseen compositional tasks, using a modular double exponential task as a case study. We demonstrated that an in-context curriculum enables zero-shot inference on compositional tasks and increases robustness. As an initial step to understand the model’s internal workings, we used a linear probe to explore how the model processes the curriculum. We found that the internal representation encodes subtask information from the curriculum blocks and the intermediate values of the compositional computation are effectively decoded as the compositional task sequence is processed. Finally, we observed that the amount of compositional task information provided in-context (controlled by curriculum length) affects both the learning strategy and the evolution of task representations during training. Our observations suggest that data properties present in context, such as curricula, can enable compositional generalization in models' emergent in-context learning abilities. 

\paragraph{Importance of structure in data} The types of compositional context structures we have emphasized in this work occur frequently in natural language data; from textbooks to novels, many documents introduce simpler elements in the beginning that build to yield more complex interactions later. Thus, while many theoretical works on in-context learning focus on presenting IID examples of a single task in context, our work highlights that language models may yield qualitatively different types of in-context learning when the contexts have a curricular compositional structure. Furthermore, our observation of diverse and even mixed strategies emerging from different curricula suggests rich inner working of in-context learning modulated by different data structures. These findings therefore highlight the importance of considering the many types of context structures that may contribute to in-context learning \citep[cf.][]{lampinen2024broaderspectrumincontextlearning}. We hope that our results will encourage more exploration of curricula and compositional learning in-context, both in controlled settings and at scale.

\paragraph{Mixtures of strategies and spectrum-like property of compositional generalization}
In Section~\ref{results_which}, we observe different curriculum designs lead to the models showing signs of both compositional and non-compositional strategies. Indeed, compositional generalization seems to be not a binary strategy choice but rather a spectrum-like behavior, echoing similar observations in natural language learning \citep[e.g.][]{rabovsky2020quasi,mcclelland2015capturing}. This suggests a complex interplay between the data structure providing compositional information and the degree of resulting compositional generalization. For example, even when the underlying compositional task structure is the same, depending on precisely how the subtask and compositional task examples are given, different levels of compositional generalization ability can be induced. We also observe that the strategy of the model is linked to the order in which the tasks are learned, which highlights the importance of dynamical aspects of the emergence of in-context learning \citep{singh2024needs, singh2024transient, park2024competition, singh2025strategy,yin2025attention}.

\paragraph{Limitations and future directions} Our analysis of mechanisms in this paper is limited to correlational evidence via linear probes. Further analysis with causal manipulation (e.g., path patching) would be necessary to gain a more precise understanding of the mechanisms behind the observed model behavior. Furthermore, in naturalistic settings there may be many types of compositional tasks that share common components, as well as cases in which the components are not always presented together with the full task; examining how our findings change in such settings would be an exciting direction for future work. It would be interesting to explore the representations of large language models as they learn novel tasks from compositional in-context curricula. The present work aims to lay the foundations for such explorations, by considering a controlled setting where we could train models from scratch with controlled manipulations of data properties.

\paragraph{Societal Impact and Reproducibility} While our work aims fundamental understanding of language models, we do not anticipate any immediate societal impact from this research. The codebase will be open-sourced upon acceptance.

\section*{Acknowledgements}
We thank Basile Confavreaux, Sara Dragutinović, Yedi Zhang for helpful discussions and feedback.
This work was supported by the following funding sources: Gatsby Charitable Foundation (GAT3850) and Sainsbury Wellcome Centre Core Grant from Wellocme Trust (219627/Z/19/Z) to JHL, AKS and AMS;
Schmidt Science Polymath Award to AMS. AMS is a CIFAR
Azrieli Global Scholar in the Learning in Machines \& Brains
program.

\bibliography{neurips2025_conference}
\bibliographystyle{plainnat}

\newpage
\appendix

\begin{figure}[H]
\section{Loss and performance curve}
\justifying

\label{app:example_loss}
    \centering
    \begin{subfigure}{0.5\linewidth} 
        \includegraphics[width=\linewidth]{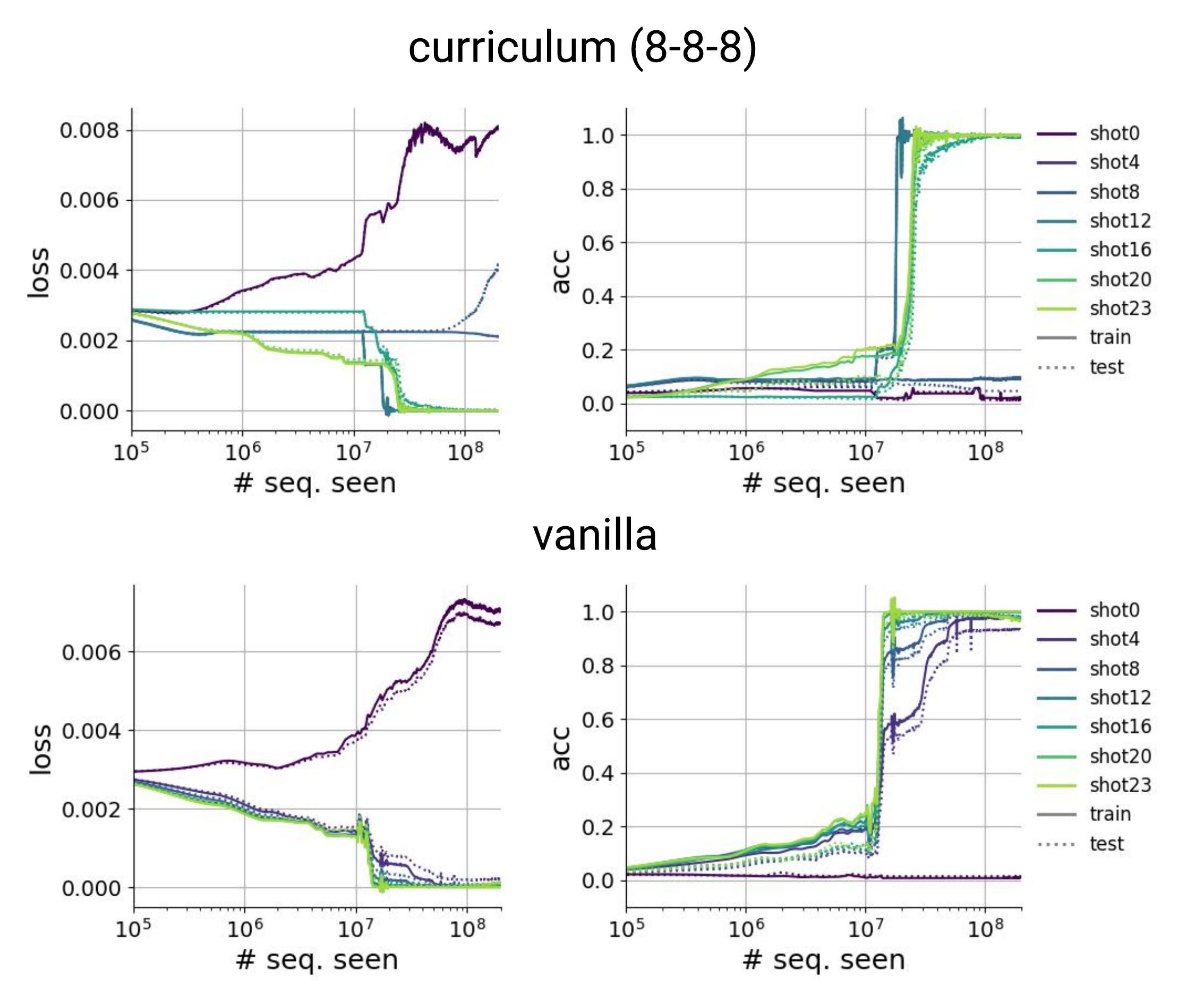} 
        \phantomcaption\label{fig:p37}
    \end{subfigure}
    \caption{Example loss and performance evolution on the curriculum (8-8-8) model and the vanilla model. In curriculum model, we observe the loss at all shots go down except for the shot 0 and shot 8, which correspond to zero-shot inference of two single tasks, since there is no information available of the task. Furthermore, we observe loss at the later shots in the same task sequence decreases earlier since more information is available in-context. We observe similar for the shot 0 in vanilla model. We also observe multiple plateaus followed by sudden drop of loss in both settings.}
    \label{app:exmaple_loss}
\end{figure}

\section{Extended robustness results}

\begin{figure}[H]
\subsection{Additional visualization of robustness}
\justifying
\label{app:error_viz}
    \centering
    \begin{subfigure}{\linewidth} 
        \includegraphics[width=\linewidth]{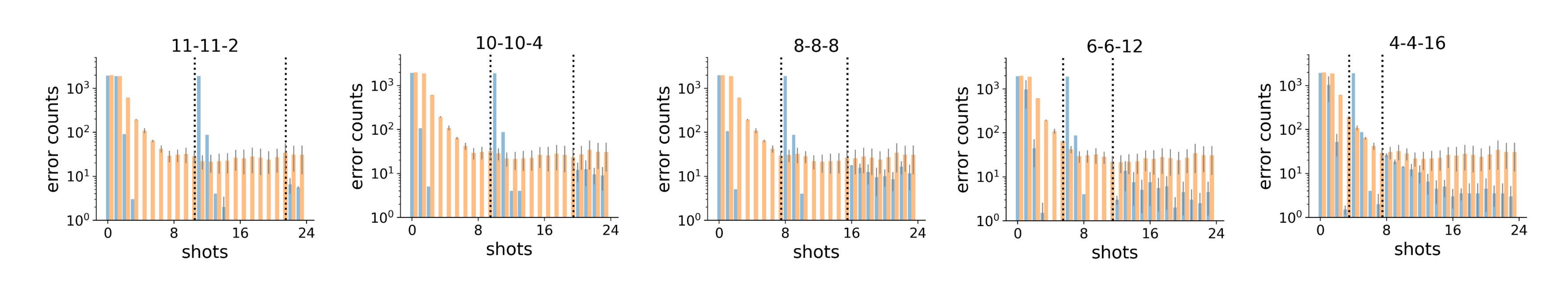} \phantomcaption\label{fig:p59_all_error}
    \end{subfigure}
    \begin{subfigure}{\linewidth} 
        \includegraphics[width=\linewidth]{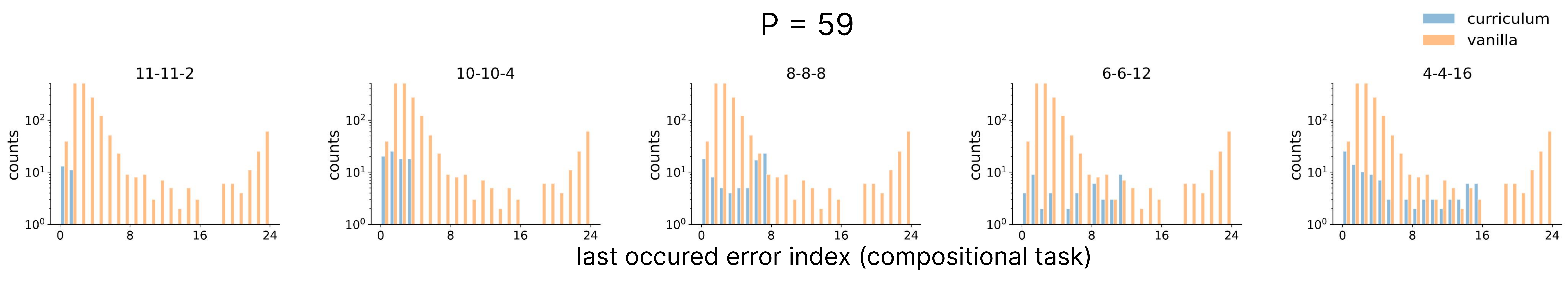} \phantomcaption\label{fig:p59_last_error}
    \end{subfigure}
    \caption{Additional visualization of error counts complementary to Figure~\ref{figure_errors_all}-\ref{figure_robustness}. \textbf{(Top)}~Error counts in entire context length across different curriculum designs. \textbf{(Bottom)}~ The counts of the last error occurred at each shot in the compositional task. We observe that the curriculum setting makes fewer errors in total compare to the vanilla models (total counts of each color) reflecting higher robustness of the curriculum models. Furthermore, strong right skewedness of the vanilla model indicates that the model tends to make errors even after many examples suggesting the model is uncertain about the task information.}
    \label{app:fig_other_modulo_all_errors}
\end{figure}

\begin{figure}[H]
\subsection{Robustness results on other values of $P$}
\justifying
We extend our results in robustness in other modulo values, $P=37$ and $P=41$. We observe that in-context curriculum can increase the robustness in compositional task generalization. The model architecture and the training setup was identical as given in Section~\ref{experimental_setup}.

All curricula designs increase robustness for $P=37$ similar to $P=59$. For $P=41$, only the curriculum (11-11-2) increases the robustness of the curriculum. We notice that when the single task block length is less than 11, the model does not learn to solve the single task (see high error bars in the first two blocks even after ~10 examples). This indicates that the model needs more single task examples to identify the each subtask and this explains why the other curricula design with shorter subtask lengths are not helping compositional generalization for $P=41$. 

\label{app:other_modulo_robustness}
    \centering
    \begin{subfigure}{\linewidth} 
        \includegraphics[width=\linewidth]{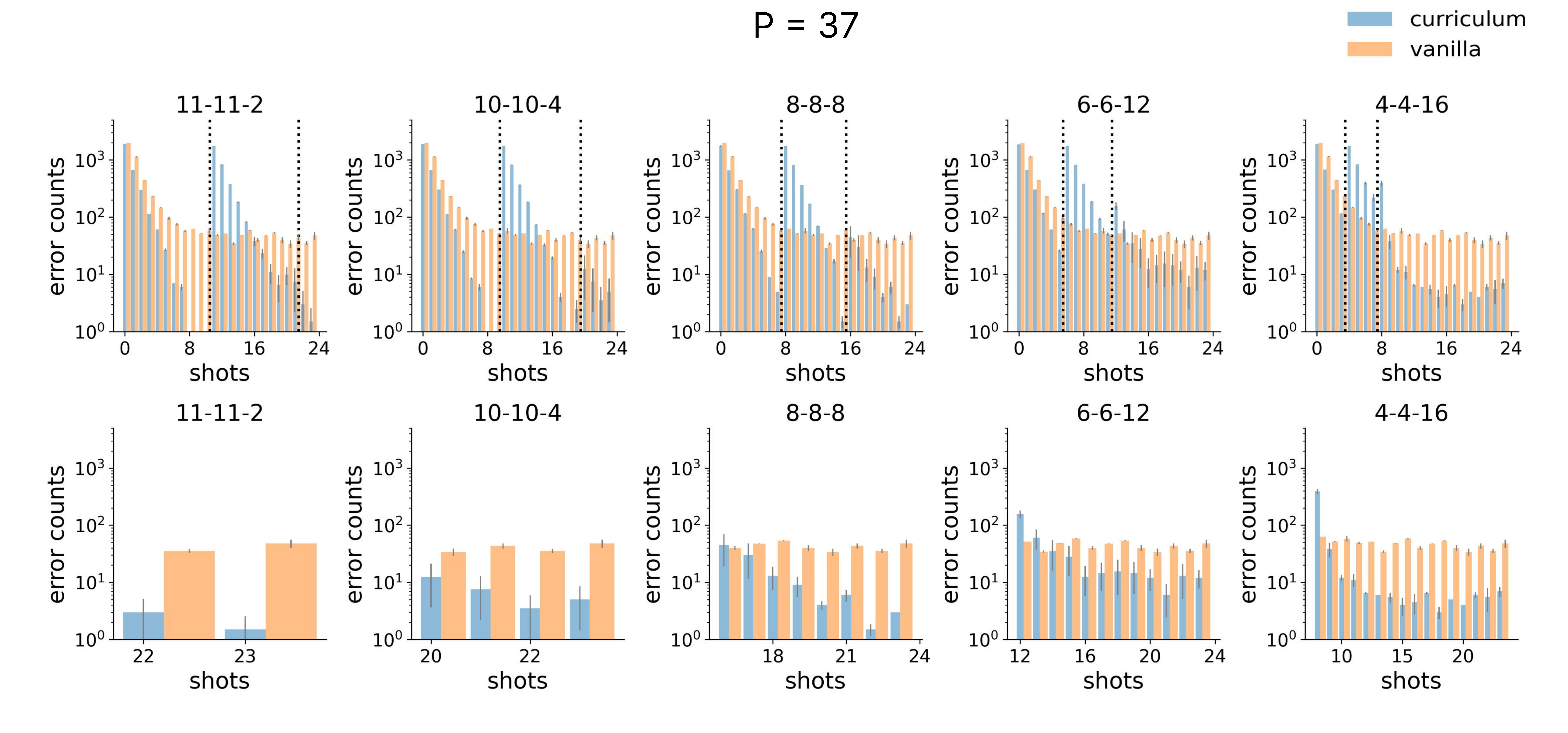} 
        \phantomcaption\label{fig:p37}
    \end{subfigure}
    \begin{subfigure}{\linewidth} 
        \includegraphics[width=\linewidth]{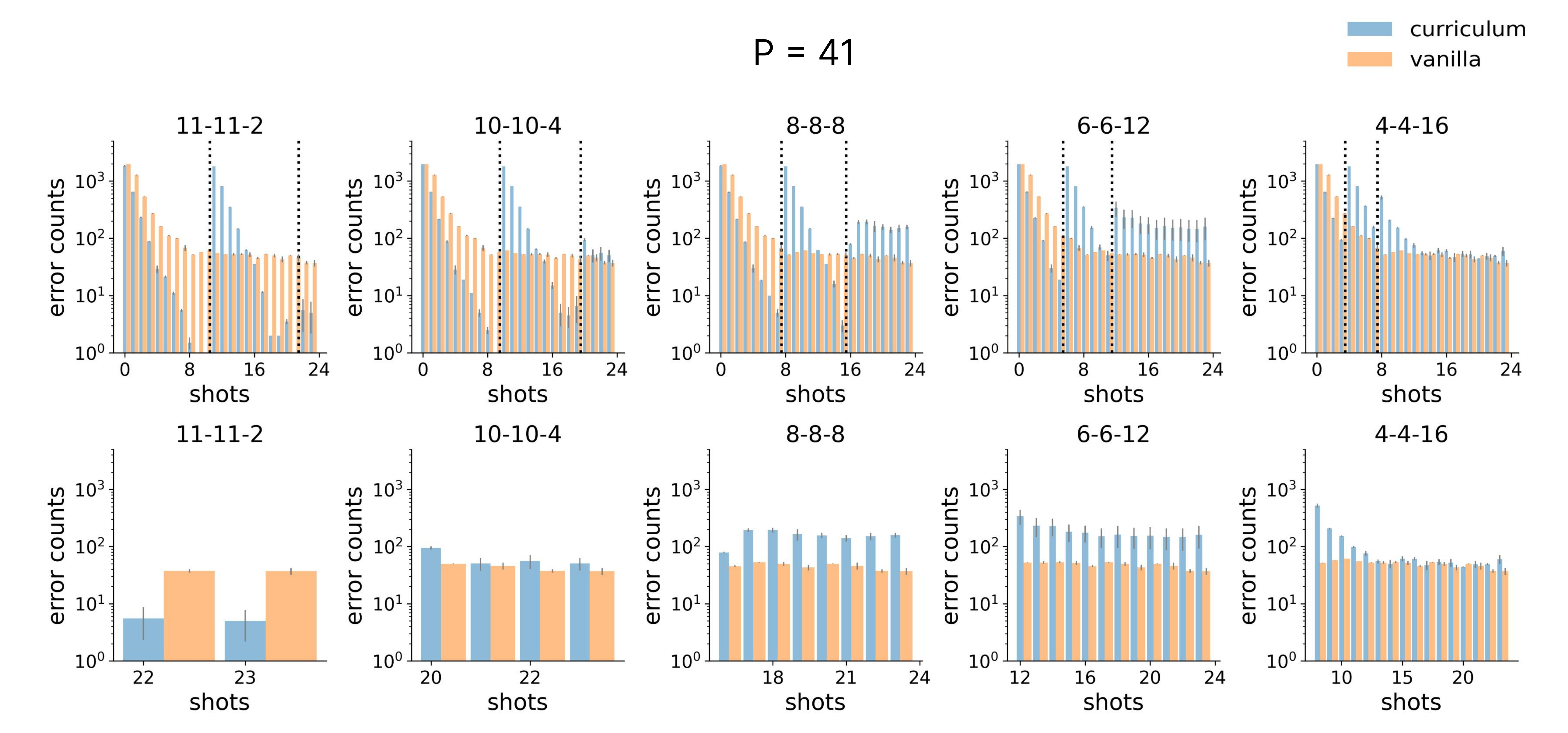} 
        \phantomcaption\label{fig:p41}
    \end{subfigure}
    \caption{Error counts in vanilla vs. curriculum model for $P=37$ and $P=41$. \textbf{(Top)}~Error counts across entire context length. \textbf{(Bottom)}~ Error counts in the compositional task block, left aligned to zero-shot at the compositional task.} 
    \label{app:fig_other_modulo_all_errors}
\end{figure}

\section{Extended results on linear probe}
\label{app:linear_probe}
We train linear probe for 1) corresponding task $y$ at each position, 2) task computation of $a$ and 3) task parameter $b$, which are required for the compositional computation of $b^{a^x}$. Since $b^{a^x} \mod P = b^{a^x \mod (P-1)} \mod P$, the intermediate values from task $a,b$ that we try to decode from compositional task blocks are $a^x \mod P$ and $b$. We train probes for the intermediate values in the subtask block as well. In subtask $a$ block, we simply decode $a^x \mod P$ (same as target value) and in subtask $b$, we decode $b$. We used unseen 1K test sequences and use $80/20$ split for training and evaluating linear probe. We used scikit \cite{scikit-learn} package for the classifier training. 
Below diagram shows what is decoded in each block. 

\begin{figure}[H]
\subsection{Linear probe additional information}
\label{app:linear_probe_decoding_value}
    \begin{center}
    \includegraphics[width=\linewidth]{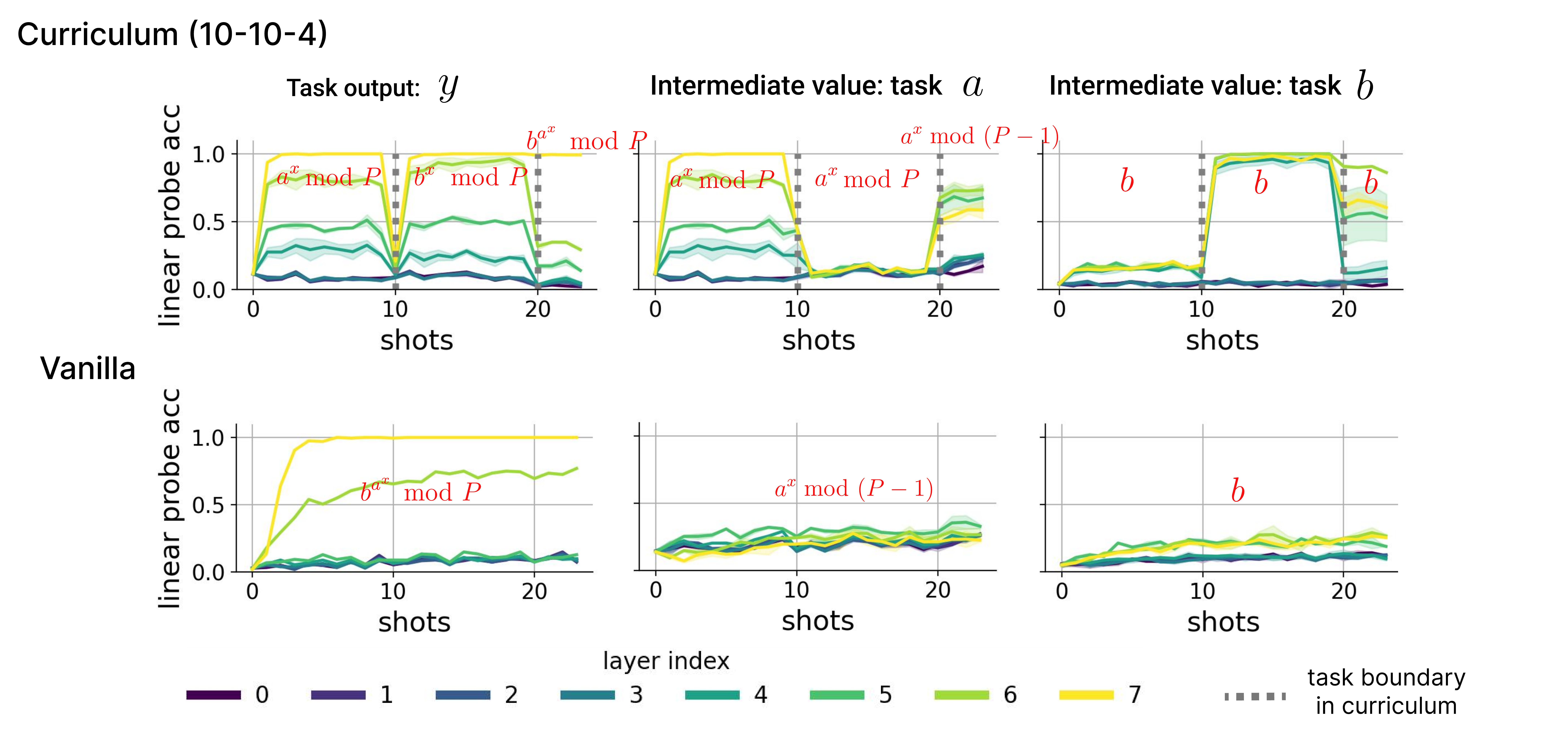}
    \end{center}
    \caption{We show what variables we are decoding in each block in main Figure~\ref{figure_probe}. Since $b^{a^x} \mod P = b^{(a^x \mod (P-1))}$, the actual intermediate computation value from task $a$ used for the compositional task is $a^x \mod (P-1)$, and we can decode this value in the compositional task block when curriculum is given while less in vanilla model. Same for $b$. }
\end{figure}

\begin{figure}[H]
\subsection{Linear probe - control baseline}
\label{app:linear_probe_baseline}
\justifying 
We performed a control experiment with shuffled task parameters $(a,b)$ to check the baseline performance and verify that our decoding accuracy is meaningful. The below figure shows that baseline decoding accuracy from shuffled task parameters is almost 0, confirming that our probe decoding accuracy is non-trivial. 
    \begin{center}
    \includegraphics[width=0.7\linewidth]{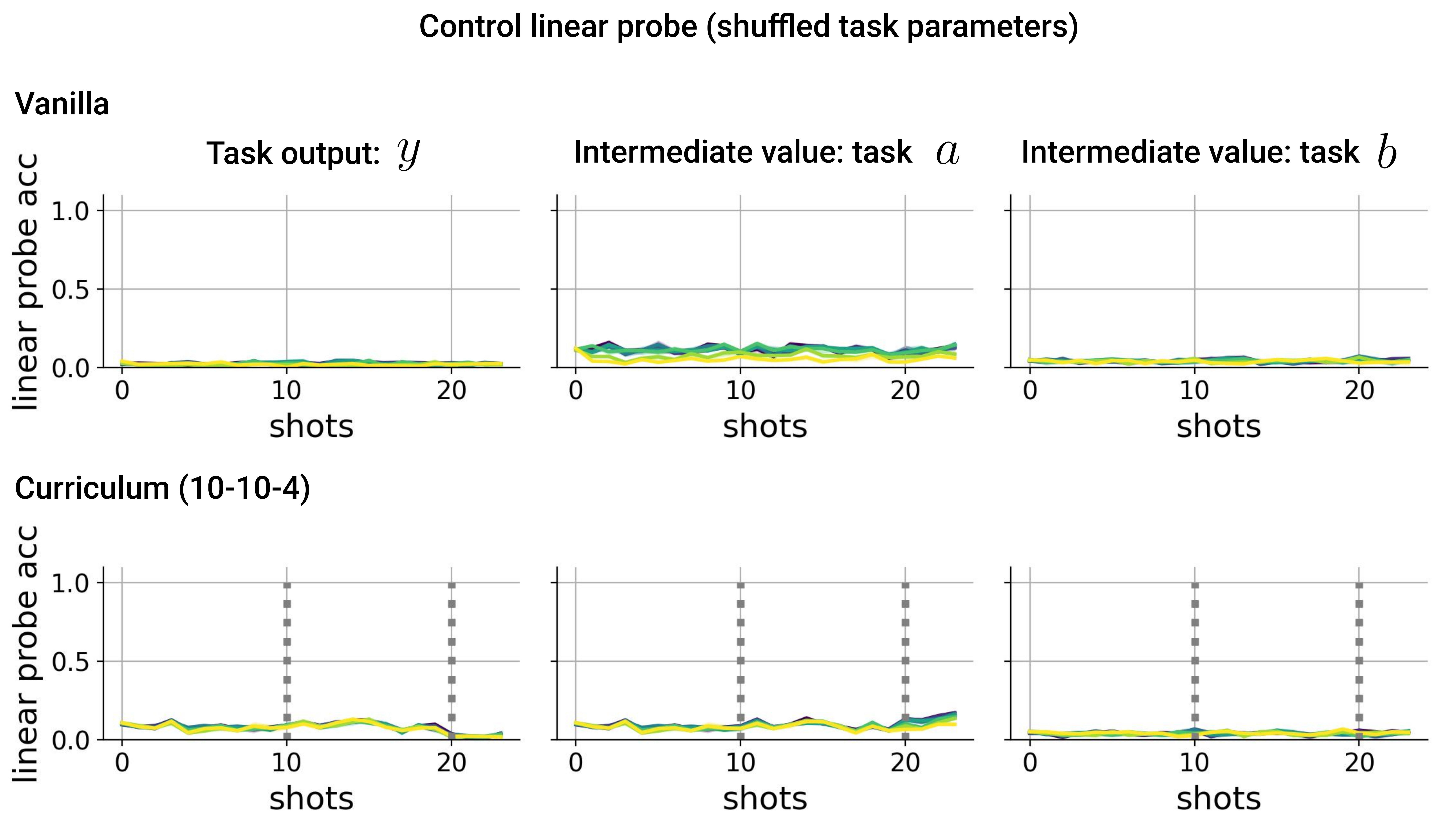}
    \end{center}
    \caption{Control linear probe decoding. We used shuffled labels for linear probe training to validate the baseline performance.}
\end{figure}

\begin{figure}[H]
\subsection{Linear probe - Varying curriculum length}
\label{app:linear_probe_diff_curr_len}
\justifying
    \begin{center}
    \includegraphics[width=\linewidth]{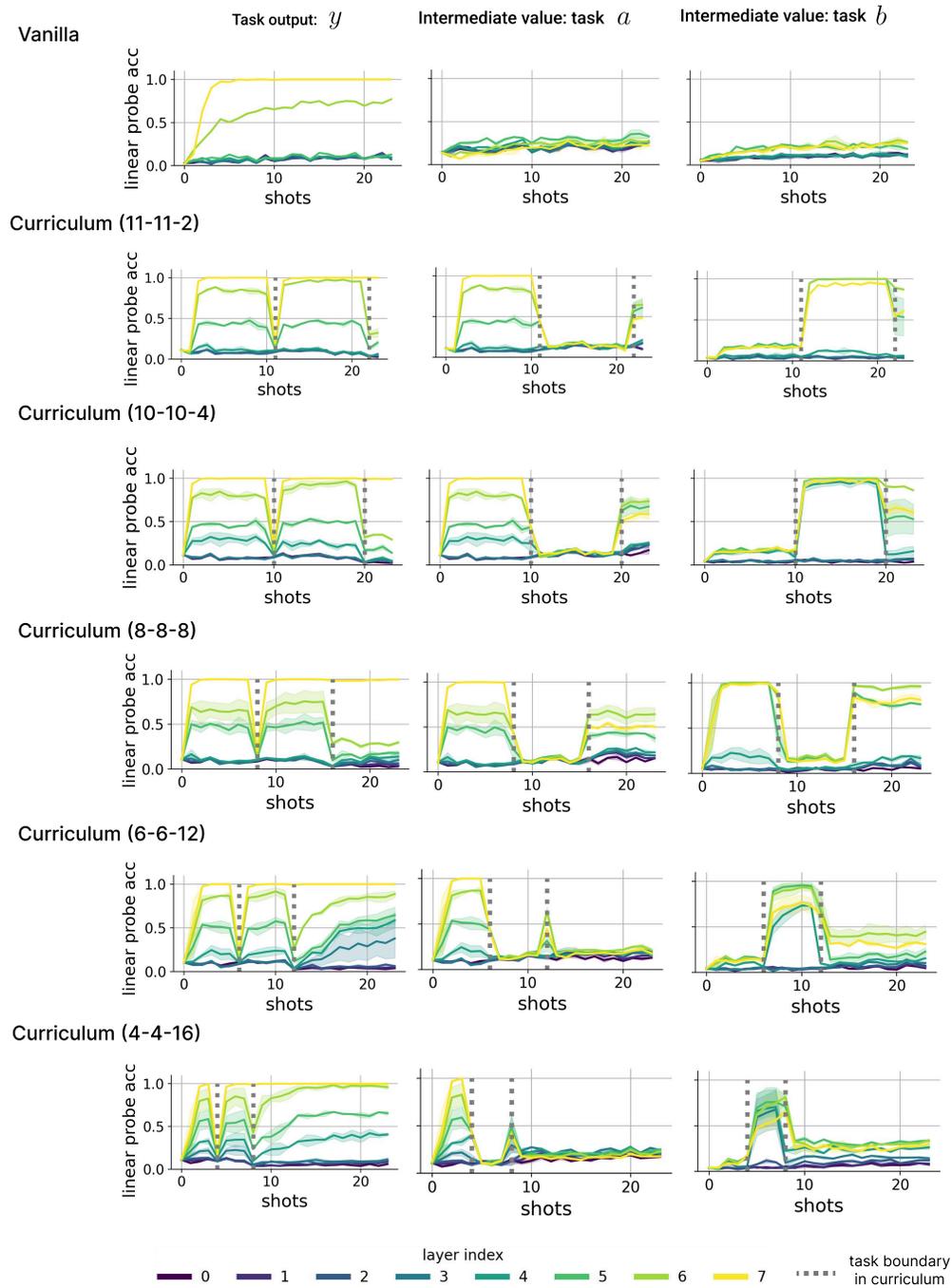}
    \end{center}
    \caption{Linear probe decoding results for other curricula design. We observe that having in-context curriculum consistently shows high decodability of intermediate values from the subtasks in compositional task block. We dive deeper into the specific high decodability at zero-shot in section~\ref{results_which}- \ref{results_when} }
\end{figure}

\begin{figure}[H]
\section{Attention pattern analysis}
\label{app:attention}
\justifying 
We visualize subset of the attention heads in layer 4-7 in the curriculum trained model and vanilla trained model, averaged on 2K evaluation sequences. In vanilla trained model, the attention pattern is continuous without outstanding block structure. On the other hand, curriculum trained model develops attention heads that show attention pattern from compositional task block to curriculum block which supports our hypothesis that the curriculum subtask representation is utilized in compositional task.
   
\begin{center}
        \includegraphics[width=\linewidth]{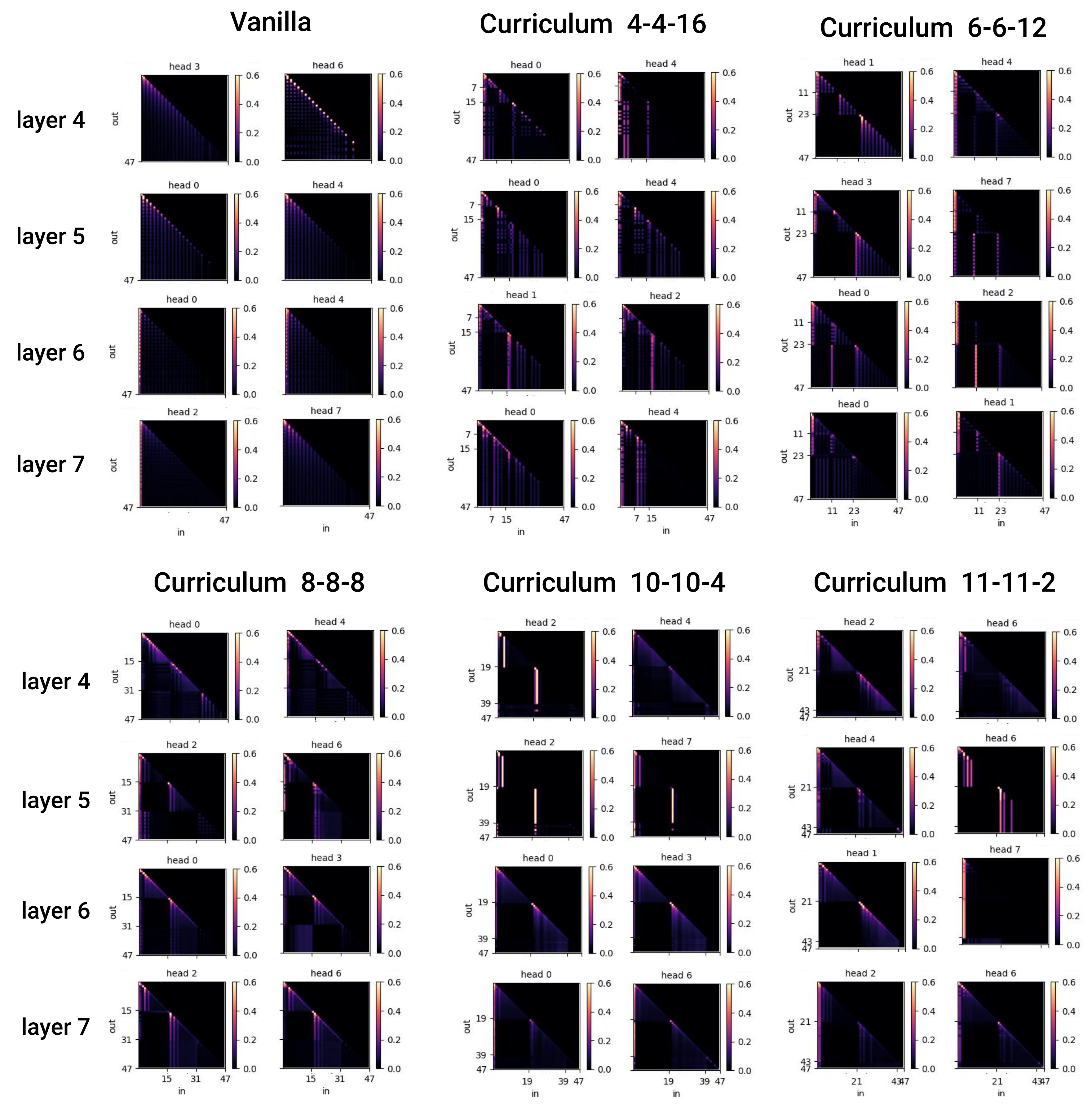}
        \caption{Attention pattern from selected heads in layer 4-7. x-axis shows token position \textit{attending from} and y-axis shows token position \textit{attending to}. We show all 48 token positions of 24 input-label pairs. For curriculum model, each task boundary is marked by ticks on x and y axis. For example, head 0 and head 3 at layer 6 from curriculum (8-8-8) shows a pattern of attending from each subtask to the last compositional task block.}
        \label{fig:app_attention_select}
\end{center}
\end{figure}

\section{Extended result on mixed strategy}

\begin{figure}
\subsection{Mismatch experiment}
\label{app:mismatch}
\justifying
    \begin{center}
    \includegraphics[width=\linewidth]{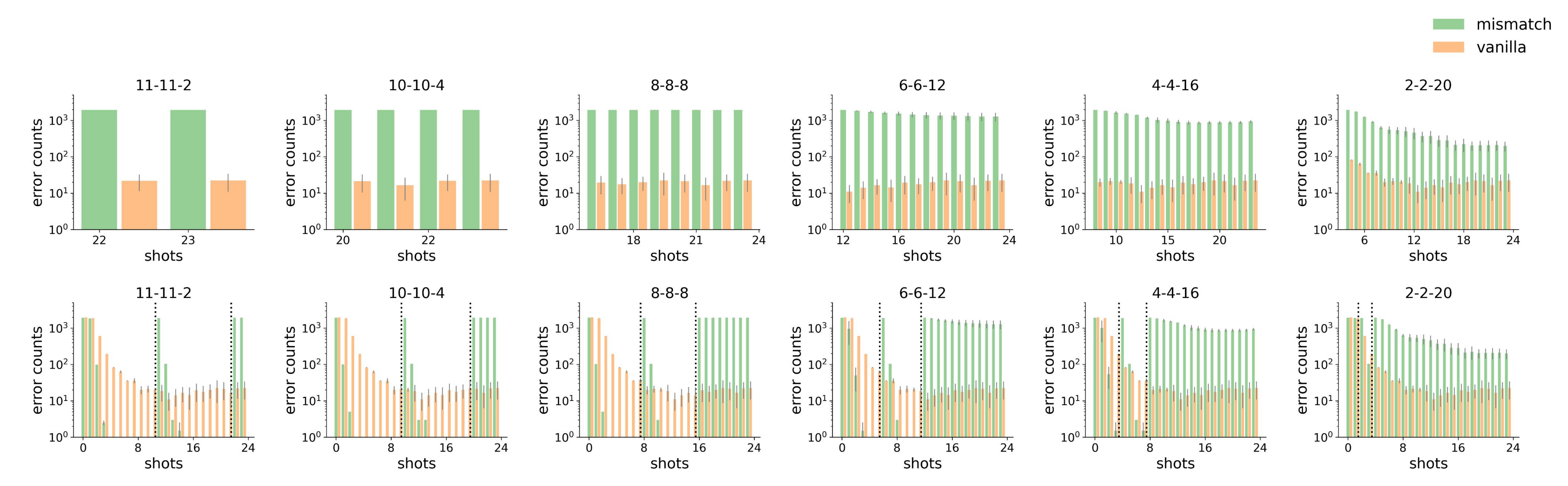}
    \end{center}
    \caption{Mismatch experiment result on other curricula designs. \textbf{(Top)}  In the curriculum with short compositional task block (11-11-2, 10-10-4, 8-8-8), we see that the model completely fails on the mismatch compositional task block. On the other hand, the curriculum with longer compositional task block (6-6-12, 4-4-16) show slight decrease as more examples are given, indicating standard few-shot learning in-context is occruing. }
\end{figure}

\begin{figure}
\subsection{Linear probe per layer}
\label{app:linear_probe_per_layer}
\justifying
    \begin{center}
    \includegraphics[width=\linewidth]{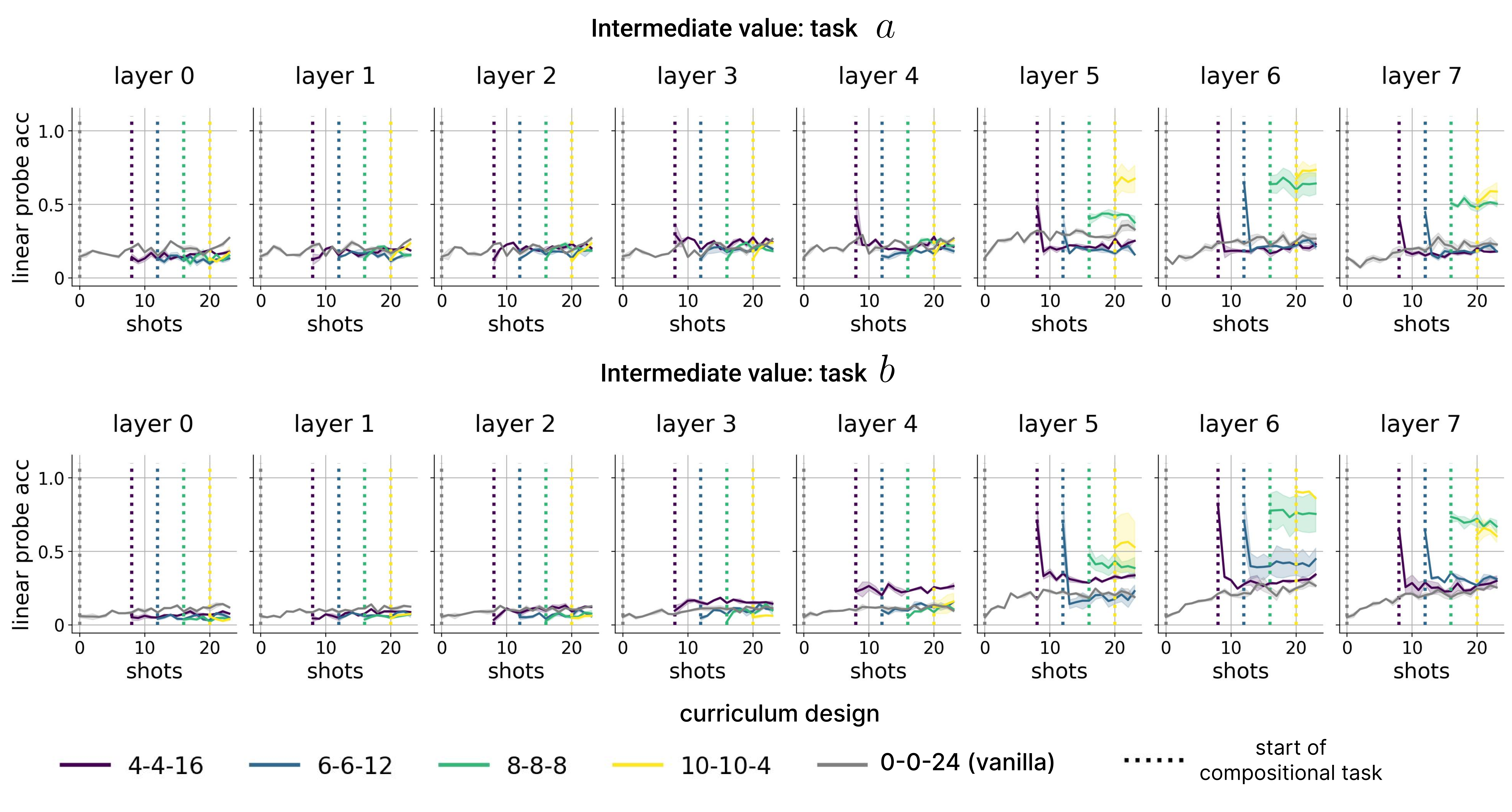}
    \end{center}
    \caption{Extended result of main Figure~\ref{figure_strategy}. We show the decoding of the intermediate values in the compositional block in all layers. }
\end{figure}

\begin{figure}
\subsection{Loss evolution over checkpoints}
\label{app:loss_all_shots}
\justifying
    \begin{center}
    \includegraphics[width=0.7\linewidth]{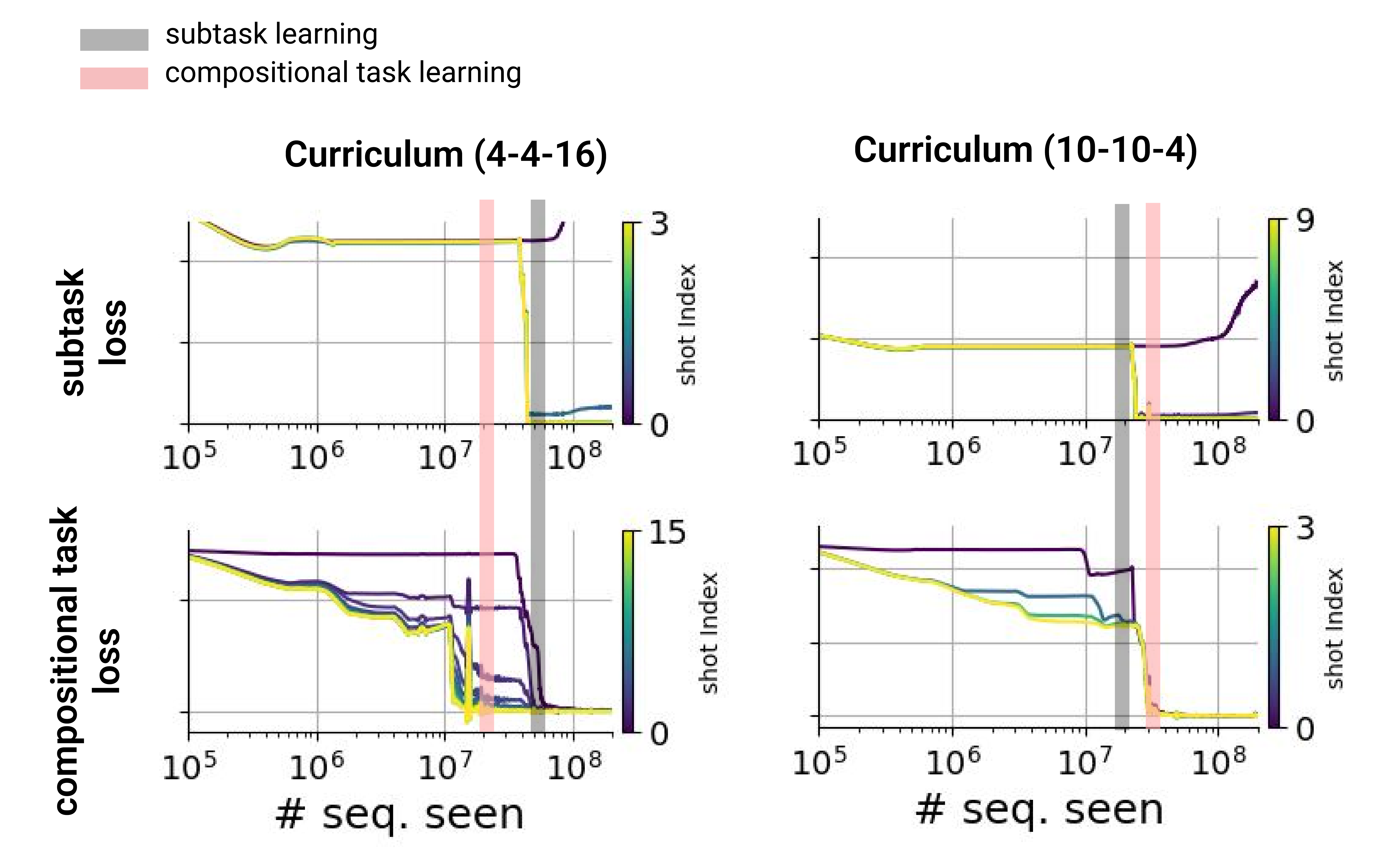}
    \end{center}
    \caption{Extended result of main Figure~\ref{figure_strategy}. We show loss curve of al shots. The loss curve above and the main Figure~\ref{figure_before_after} are filtered using Savitsky-Golay filter (using scikit implementation~\cite{scikit-learn}) with length 51 and polynomial order 3.}
\end{figure}

\begin{figure}
\subsection{Linear probing before and after the subtask learning}
\label{app:probe_before_after}
\justifying
    \begin{center}
    \includegraphics[width=\linewidth]{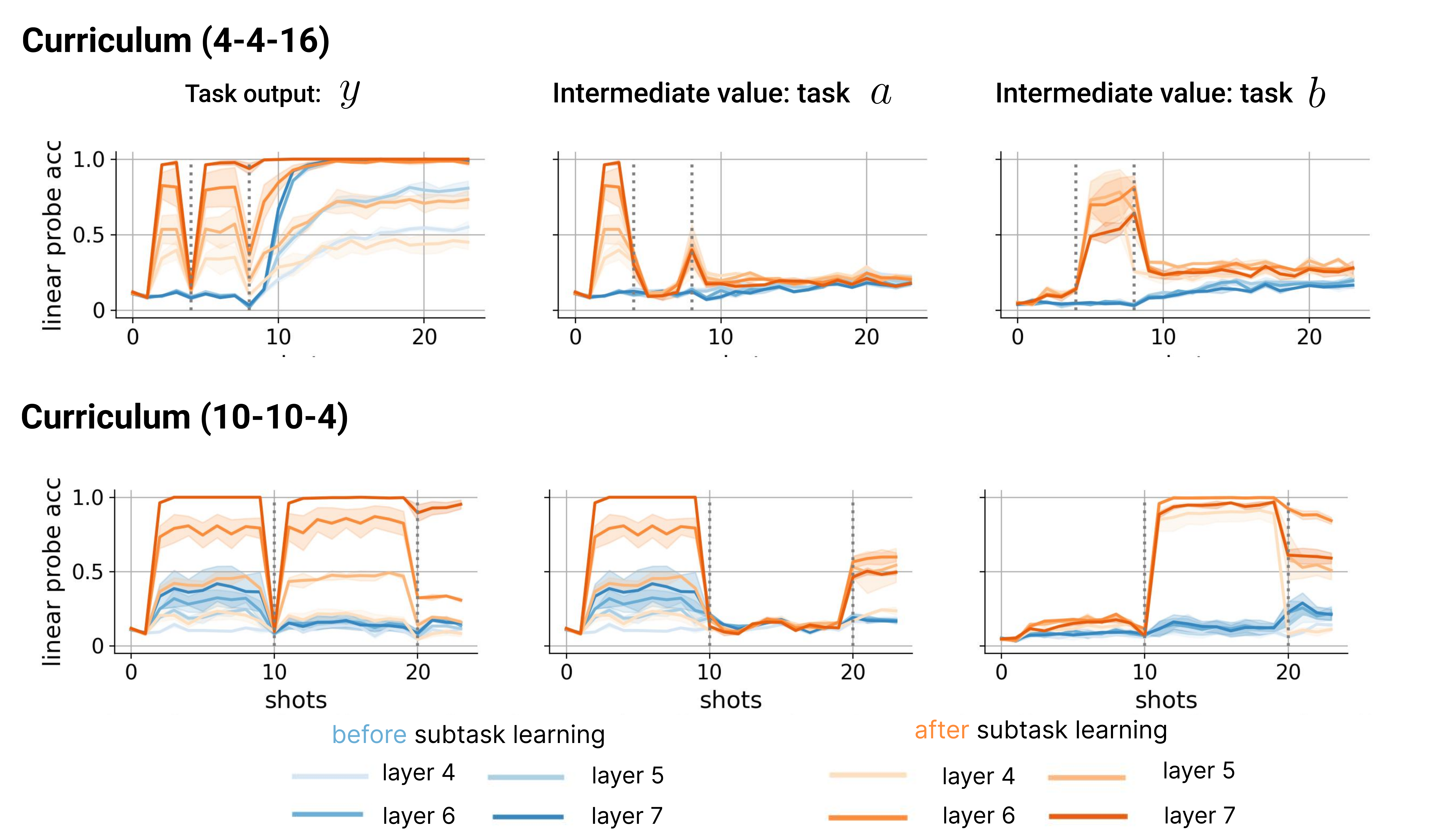}
    \end{center}
    \caption{Extended result of main Figure~\ref{figure_before_after}.}
\end{figure}

\begin{figure}[H]
\subsection{Linear probing at additional checkpoints}
\label{app:probe_more_ckpts}
\justifying
We visualize the linear probe of curriculum (10-10-4) and (4-4-16) settings at more checkpoints. We first observe that the subtask learning happens much faster in curriculum. This is because curriculum (10-10-4) contains more examples of subtask (single exponentials) compare to (4-4-16). 
    \begin{center}
    \includegraphics[width=\linewidth]{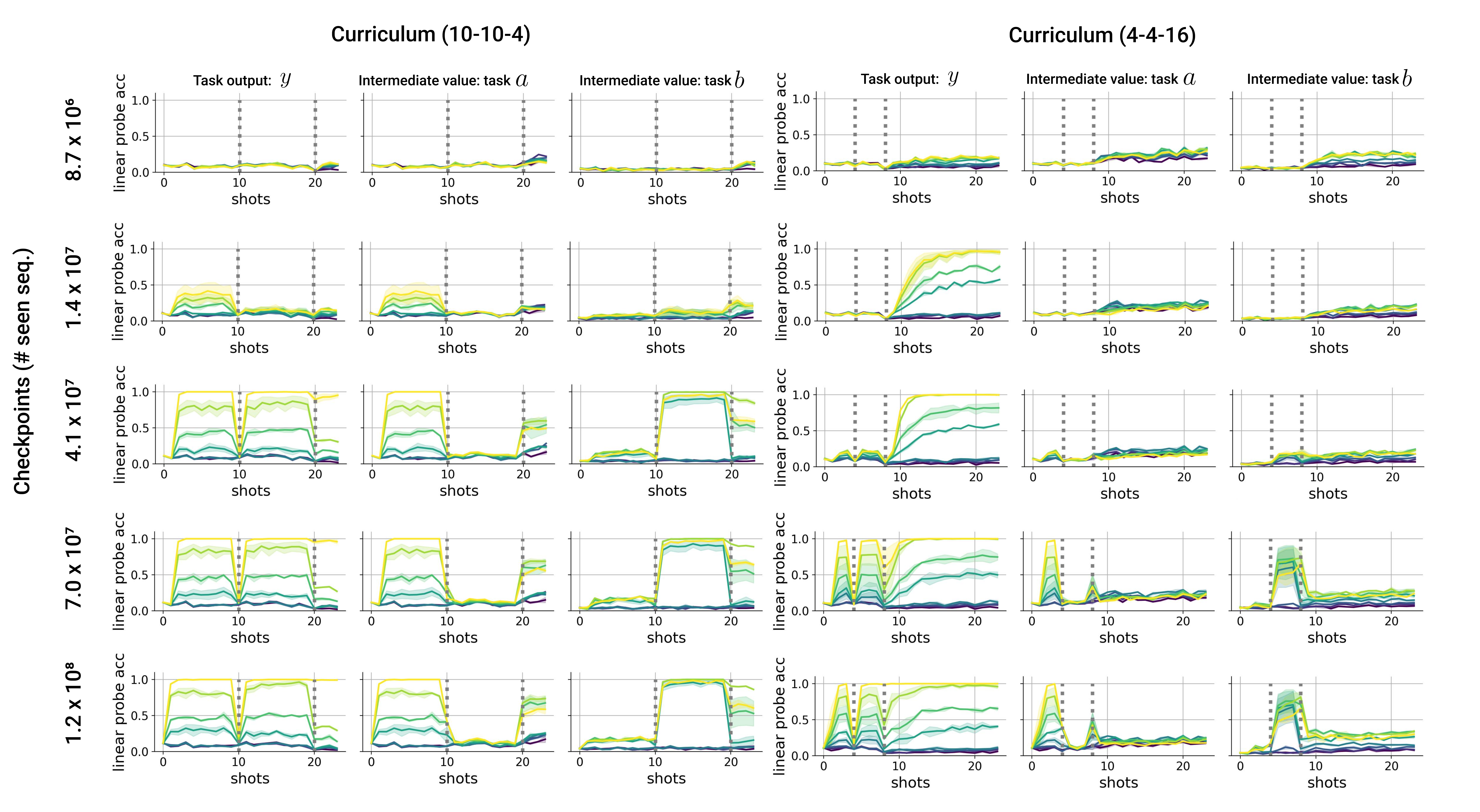}
    \end{center}
    \caption{Linear probe of linear probe of curriculum (10-10-4) and (4-4-16) settings at more checkpoints. }
\end{figure}

\section{Computing resources}
\label{app:compute}
All experiments were run on a node with 8 NVIDIA H100 GPUs and 192 CPUs. Training each model took 2–3 wall-clock hours. A set of linear probe analyses took approximately 0.5 hours. For the final analysis, we used 42 models. For the results reported in the paper, we estimate a total of roughly 105 compute hours, or 12 hours when accounting for parallelization. Including test and failed experiments, we estimate multiplication of factor $\sim5$.

\end{document}